\documentclass{article}

\usepackage[nonatbib,final]{neurips_2022}
\usepackage[numbers]{natbib}
\usepackage{wrapfig}
\usepackage[utf8]{inputenc} 
\usepackage[T1]{fontenc}    
\usepackage{hyperref}       
\usepackage{url}            
\usepackage{booktabs}       
\usepackage{amsfonts}       
\usepackage{nicefrac}       
\usepackage{microtype}      
\usepackage{xcolor}         
\usepackage{graphicx}
\usepackage{bbm}
\usepackage{paralist}
\usepackage{xspace}
\usepackage{bm}
\usepackage{multirow}
\usepackage{amsmath}

\usepackage{amssymb}
\usepackage{bbding}

\usepackage{colortbl}

\usepackage{capt-of}

\newtheorem{definition}{Definition}[section]

\newcommand{\balpha}{\bm{\alpha}}%
\newcommand{\blambda}{\bm{\lambda}}%

\newcommand\scalemath[2]{\scalebox{#1}{\mbox{\ensuremath{\displaystyle #2}}}}

\newcommand{\air}{{\small{\tt airplane}}\xspace}
\newcommand{\auto}{{\small{\tt auto}}\xspace}

\newcommand{\dep}{{\tt depth}\xspace}
\newcommand{\wid}{{\tt width}\xspace}
\newcommand{\lr}{{\tt lrn\_rate}\xspace}
\newcommand{\dr}{{\tt drop\_out}\xspace}

\newcommand{\svdd}{{\small{\sf DeepSVDD}}\xspace}
\newcommand{\rand}{{\small{\sf RandNet}}\xspace}
\newcommand{\gan}{{\small{\sf GANomaly}}\xspace}
\newcommand{\vae}{{\small{\sf VanillaAE}}\xspace}
\newcommand{\rda}{{\small{\sf RDA}}\xspace}

\newcommand{\ifor}{{\small{\sf IF}}\xspace}

\newcommand{\aes}{{\small{\sf AE-S}}\xspace}
\newcommand{\aei}{{\small{\sf AE-i}}\xspace}

\makeatletter
\newcommand\footnoteref[1]{\protected@xdef\@thefnmark{\ref{#1}}\@footnotemark}
\makeatother

\newcommand{\cbit}{\begin{compactitem}}
\newcommand{\ceit}{\end{compactitem}}
\newcommand{\cben}{\begin{compactenum}}
\newcommand{\ceen}{\end{compactenum}}

\newcommand{\beq}{\begin{equation}}
	\newcommand{\eeq}{\end{equation}}

\newcommand{\bit}{\begin{itemize}}
	\newcommand{\eit}{\end{itemize}}
\newcommand{\ben}{\begin{enumerate}}
	\newcommand{\een}{\end{enumerate}}

\newcounter{x}

\newcommand{\identity}{\mathbbm{1}}
\newcommand{\realR}{\mathbb{R}}
\newcommand{\br}{\mathbf{r}}
\newcommand{\bw}{\mathbf{W}}
\newcommand{\bss}{\mathbf{s}}
\newcommand{\bR}{\mathbf{R}}
\newcommand{\bSS}{\mathbf{S}}

\newcommand{\method}{{\sc RobOD}\xspace}
\newcommand{\imethod}{i-{\sc RobOD}\xspace}

\title{Hyperparameter Sensitivity in Deep Outlier Detection \\
{\Large{Analysis and a Scalable Hyper-Ensemble Solution}}}

\author{%
  Xueying Ding \\
  Carnegie Mellon University\\
  \texttt{xding2@cs.cmu.edu} \\
   \And
   Lingxiao Zhao\\
   Carnegie Mellon University\\
  \texttt{lingxiao@cmu.edu} \\
   \And
    Leman Akoglu \\
   Carnegie Mellon University\\
   \texttt{lakoglu@cs.cmu.edu} \\
}

\begin{document}

\maketitle

\begin{abstract}
Outlier detection (OD) 
literature exhibits numerous 
algorithms as it applies to diverse domains.
However, given a new detection task, it is unclear how to choose an algorithm to use, nor how to set its hyperparameter(s) (HPs)
in unsupervised settings. HP tuning is an ever-growing problem with the arrival of many new detectors based on deep learning, which usually come with a long list of HPs.  Surprisingly, the issue of model selection in the outlier mining literature has been ``the elephant in the room''; a significant factor in unlocking the utmost potential of deep methods, yet little said or done to systematically tackle the issue.
In the first part of this paper, we conduct the first large-scale analysis on the HP sensitivity of deep OD methods, and through more than 35,000 trained models, quantitatively demonstrate that model selection is inevitable.
Next, we design a HP-robust and scalable deep 
 \textit{hyper-ensemble} model called \method that assembles models with varying HP configurations, bypassing the choice paralysis.
 Importantly, 
  we introduce novel 
strategies to speed up ensemble training, such as
parameter sharing, batch/simultaneous training, and data subsampling, that allow us to
train fewer models with fewer parameters.
Extensive experiments on both image and tabular datasets show that \method achieves and retains robust, state-of-the-art detection performance as compared to its modern counterparts, while taking only $2$-$10$\% of the time by the na\"ive hyper-ensemble with independent training. 
\end{abstract}

\section{Introduction}
\label{sec:intro}


Outlier detection (OD) finds numerous real-world applications in finance, security, healthcare, to name a few. Thanks to this popularity, the literature has grown to offer a large catalog of detection algorithms \cite{bookCharu}.
With the recent advances in deep learning, the literature has been booming with the addition of many more OD models based on deep neural networks (NNs). (See surveys \cite{chalapathy2019learning,pang2021deep,UnifyingSurvey}.)

While there is no shortage of OD methods today, given a new task, it is unclear how to choose {which algorithm or model} to use, nor {how to configure} its hyperparameter(s) (HPs) in unsupervised settings. That is, the fundamental problem of outlier model selection remains vastly understudied.
Several evaluation studies have illustrated the sensitivity to HPs 
for traditional (i.e. non-deep) OD methods  \cite{journals/sigkdd/AggarwalS15,DBLP:journals/datamine/CamposZSCMSAH16,goldstein2016comparative}. 
Most surprisingly, however, the issue of HP tuning/model selection for the newly burgeoning deep OD models has been ``the elephant in the room''; a well-known problem that no one seems to want to bring up, which is exactly the focus of this paper.

Deep OD models are promising thanks to appealing properties such as task-driven representation learning and end-to-end optimization. On the other hand, while their traditional counterparts had only 1-2 HPs\footnote{\label{note1}e.g., $k$ in nearest neighbor based 
	LOF \cite{breunig2000lof}, $\nu$ in OCSVM \cite{scholkopf2000support}, sample size $\psi$ and \#trees $t$ in IF \cite{liu2008isolation}.}, deep OD models come with a long list of HPs: 
($i$) 
architecture HPs (e.g. depth, width), ($ii$) regularization HPs (e.g. dropout, weight decay rates),
and ($iii$) optimization HPs (e.g. learning rate, 
epochs). 
These are 
inherited ones from regular deep NNs while 
most also exhibit their model-specific/specialized HPs. It would not be a 
freak occurrence to assume
that their performance is heavily dependent on these HP settings.
However, our closer analysis of the experiment testbeds in recent literature on deep OD models falls far from systematically addressing the issue. (See Appx. \ref{ssec:existing} 
Table \ref{tab:unsupmethods} for a preview.)
We find hardly any discussion on model selection, with only a few work empirically studying sensitivity but to model-specific HPs only.
Majority of work report results for a single ``recommended'' (how, unclear) configuration used for all datasets, or tune only a subset of the HPs on labeled validation and sometimes even test (!) data.
To the best of our knowledge, there is no existing work that attempts (unsupervised) model selection for deep OD models. 




In this work, our research goals are two-fold.
\textbf{First}, 
through extensive experiments,
we quantitatively demonstrate that deep OD models from various families  are all sensitive to their HP settings. 
Our analysis shows that model selection is inevitable and 
 is key to truly unlock the utmost potential of deep OD models.
\textbf{Second}, motivated by our analysis, we propose a scalable deep 
\textit{hyper-ensemble} called \method that
obviates HP selection 
through assembly 
of deep autoencoders 
with varying HP configurations.
To speed up ensemble training, 
we introduce novel architectural and training strategies, and train fewer models, with fewer parameters, on smaller subsamples of data;
by leveraging parameter sharing and joint/simultaneous training. 
The main contributions of our work are 
as follows.


\vspace{-0.05in}
\cbit
\item {\bf First large-scale study on HP sensitivity of deep OD models:~} 
We build a large testbed to systematically measure the performance variability of deep OD models under varying HP settings. Our study involves models from four different families, 
on both image and vector data, under both ``clean'' (i.e. inlier only) as well as ``polluted'' (i.e. outlier-contaminated) training data, over 3 random initializations, 80-800+ different HP configurations per model across 4-8 unique HPs. Overall, our analysis involves more than 35,000 runs. (Sec. \ref{sec:sensitive})

\item {\bf \method, a new deep hyper-ensemble OD model:~} 
Motivated by our empirical study, we propose a hyper-ensemble model called \method which combines scores from a collection of models, each trained with a different HP configuration. Rather than trying to choose, \method fully bypasses the choice of and hence sensitivity to HP settings, and achieves robust (i.e. stable) performance across different initializations. (Sec. \ref{ssec:overview})


\item {\bf Design strategies to speed up hyper-ensemble training:~} 
We propose speed-up strategies to efficiently hyper-ensemble model depth and width. 
We use an autoencoder (AE) with skip-connections to 
simultaneously train multiple AEs with different depths. In addition, we employ batch 
training of multiple models and use zero-masking on shared parameters
 to get different widths. 
Together, these provide a {90-98$\%$ savings in running time}. 
(Sec. \ref{ssec:design})

\item {\bf Extensive experiments:~} Besides our large-scale measurement study, we also perform experiments on additional 
benchmark datasets, comparing \method to baseline deep OD models as well as a traditional 
tree-ensemble. 
\method achieves competitive or often better performance which, importantly, exhibits low variance by random initialization. 
(Sec. \ref{sec:experiments})
\ceit
\vspace{-0.05in}

We expect that our work will increase awareness and help shift the community's focus (at least to some extent) from building the next yet-another deep OD model toward the fundamental issue of unsupervised model selection and hyperparameter-robust model design.
To foster future research, we open source all code and datasets at {{\url{https://github.com/xyvivian/ROBOD}}}.


%

\section{Related Work}
\label{sec:related}


\textbf{Unsupervised Outlier Detection (OD).~}
There exists a large pool of what-is-now-called traditional, i.e. not deep learning based, OD methods \cite{bookCharu,chandola2009anomaly}. 
These methods work with the original features or subspaces thereof, and typically exhibit just one or two hyperparameters (HPs).\footnoteref{note1}
With recent advances in deep learning, there has been a boom in deep OD models, as those can learn new task-dependent feature representations and directly optimize an OD objective. Despite their short history, multiple surveys have been published that aim to cover this fast-growing literature \cite{chalapathy2019learning,pang2021deep,UnifyingSurvey}.
While deep OD models have been shown to outperform their traditional counterparts, they exhibit a much longer list of (typically 4-8) HPs (e.g. depth, width, dropout, weight decay, learning rate, epochs, etc. besides other model-specific HPs)  that makes them very challenging to tune in unsupervised settings.

\textbf{\bf Model Selection in Unsupervised OD: Prior (Black) Art.~}
At large, unsupervised outlier model selection remains to be a vastly understudied,
yet extremely important area.
 Various evaluation studies have reported 
 traditional detectors 
 to be quite
sensitive to their HP choices
\cite{journals/sigkdd/AggarwalS15,DBLP:journals/datamine/CamposZSCMSAH16,goldstein2016comparative}, raising concern for the fair evaluation and comparison of different models.
Earlier work on automatically selecting HPs are limited to one-class models \cite{ghafoori2018efficient,tax2001outliers,wang2018hyperparameter}.
More recently, general-purpose \textit{internal} (i.e., unsupervised)  model evaluation heuristics have been proposed \cite{journals/corr/Goix16,MarquesCZS15,JCC8455},
which solely rely on the input data (without labels) and the output (i.e., outlier scores). 
MetaOD \cite{zhao2021automatic} employs meta-learning to transfer information from similar historical tasks to a new task for model selection, which 
has only been tested on traditional OD models.
Different from those that aim to select a single model, ensemble models have also been employed for OD \cite{aggarwal2017outlier}, including those that combine models from the same family 
\cite{conf/kdd/LazarevicK05} as well as heterogeneous
detectors from different families \cite{journals/tkdd/RayanaA16}.

%


Regarding deep OD methods,
we have surveyed a large collection of recent papers 
and their experimental testbed and HP settings, a summary of which is given in Appx. \ref{ssec:existing} Table \ref{tab:unsupmethods}.
To our surprise, we found \textit{hardly any discussion on model selection}, with only a few work presenting sensitivity analysis with respect to not all but some, model-specific HPs.  
While some work reserve labeled validation/hold-out data to tune a subset of the HPs \cite{boyd2020neural,goyal2020drocc,hu2020hrn},
majority of them fix the HP values and call them ``recommended''/default settings \cite{bergman2020classificationbased,conf/sdm/ChenSAT17,conf/ipmi/SchleglSWSL17,journals/mia/SchleglSWLS19,zenati2018efficient,conf/iclr/ZongSMCLCC18}. 
Moreover, a non-negligible number of existing work choose some critical HPs empirically on test data (!) to yield optimum results \cite{akcay2018ganomaly,conf/icml/RuffGDSVBMK18,conf/kdd/ZhouP17} (See Table \ref{tab:unsupmethods}, last column).  
Some work that builds on previous models (e.g., deep SVDD-based methods \cite{conf/icml/RuffGDSVBMK18} vs. multi-sphere extension \cite{ghafoori2020deep}, transformation-based methods 
\cite{conf/nips/GolanE18} for images
vs. their extension to vector data
\cite{bergman2020classificationbased}, AnoGAN \cite{conf/ipmi/SchleglSWSL17} and the follow-up EGBAD \cite{zenati2018efficient}) use the same architecture and HP settings as the prior work for consistent/``fair'' comparison. However it is unlikely that the same HP values would work comparably for different models. 

Admittedly, it is challenging to tune (a long list of) HPs in the absence of labels, yet, 
the opacity 
in the deep OD literature 
warrants careful investigation on the stability of model performance under varying HP settings, and ultimately on the fair comparison between these and traditional OD methods.




\textbf{Deep Model Ensembles.~} Recently, 
deep NN predictions have been found to be often poorly calibrated \cite{guo2017calibration}. 
As Bayesian learning does not offer straightforward training, 
deep ensemble models 
have been proposed as a simple alternative \cite{conf/nips/Lakshminarayanan17} to improve 
predictive uncertainty,  
as well as efficient ways of training deep NN ensembles 
\cite{conf/mlsys/GuanMSLP20,conf/iclr/WenTB20}. 
In this work, we leverage ensemble modeling toward a different goal: to improve the stability and robustness of unsupervised OD models to HP settings, combining predictions from models with different HPs into an OD 
\textit{hyper}-ensemble. 
The closest to our work is Wenzel \textit{et al.}'s deep hyper-ensemble
\cite{wenzel2020hyperparameter}, which, different from ours, considers {\em supervised} problems, 
to further foster diversity in the ensemble and thereby achieve better uncertainty estimation.

%


\section{Hyperparameter-Sensitivity Analysis of Deep OD}
\label{sec:sensitive}

\vspace{-0.05in}
\subsection{Testbed Setup}
\label{ssec:setup}
\vspace{-0.05in}


{\bf Models.} We study HP sensitivity of five deep OD methods of four different types: a basic deep autoencoder \vae trained with reconstruction loss, \textit{robust} deep autoencoder \rda \cite{conf/kdd/ZhouP17}, \textit{one-class} classification based \svdd \cite{conf/icml/RuffGDSVBMK18}, \textit{adversarial} training based \gan \cite{akcay2018ganomaly}, and 
an (AE) \textit{ensemble} model \rand \cite{conf/sdm/ChenSAT17}. These exhibit 4 to 8 HPs, as listed in Table \ref{tab:hps}. (See Appx. \ref{ssec:hps} for descriptions.) (Note that \rand is \textbf{not} a \textbf{hyper}-ensemble: members use the same HP configs except for NN sparsity.)
We define a grid of 2-3 different values for each HP, including the author-recommended values when available (See details in Appx. Table \ref{tab:grids}), and train each deep OD method with all combinations; yielding 81-864 different models. (Note that 192 \rand models each consists of an ensemble of 50 or 200 AEs.)
We repeat each experiment 3 times 
with different initializations.

\begin{table}[!ht]
	\setlength{\tabcolsep}{1pt}
\caption{
Deep OD models used for studying hyperparameter sensitivity.
We give the number (in parenthesis) and the list of HPs for each method, along with the total number of models  trained for evaluation. (See Appx. \ref{ssec:hps} for HP descriptions and Appx. Table \ref{tab:grids} for list of grid values per HP.) 
\label{tab:hps}
}
\vspace{-0.05in}
\hspace{-0.05in}
{\scalebox{0.85}{
\begin{tabular}{l|lr}
\toprule
\textbf{Method}  & \textbf{List of hyperparameters (HPs)}  & \textbf{\#models}   \\
\midrule
\vae   & (4)  n\_layers $\cdot$ layer\_decay $\cdot$ LR $\cdot$ iter  & 81 \\
\rda \cite{conf/kdd/ZhouP17}  & (6) $\lambda$ $\cdot$ n\_layers $\cdot$ layer\_decay $\cdot$ LR $\cdot$ inner\_iter $\cdot$ iter  & 324  \\
\svdd \cite{conf/icml/RuffGDSVBMK18}  & (8) conv\_dim $\cdot$ fc\_dim $\cdot$ Relu\_slope $\cdot$ pretr\_iter $\cdot$ pretr\_LR $\cdot$ iter $\cdot$ LR $\cdot$ wght\_dc  & 864   \\
\gan \cite{akcay2018ganomaly}  &  (6) $w_{adv}=1$ $\cdot$ $w_{con}$ $\cdot$ $w_{enc}$ $\cdot$ z\_dim $\cdot$ LR $\cdot$ iter & 162 \\
\rand \cite{conf/sdm/ChenSAT17}   & (8) n\_layers $\cdot$ layer\_decay $\cdot$ sample\_r $\cdot$
ens\_size $\cdot$ pretr\_iter$=100$ $\cdot$ iter $\cdot$ LR $\cdot$   wght\_dc$=0$  & 192 \\
 \bottomrule
\end{tabular}
}}
\vspace{-0.1in}
\end{table}


{\bf Train/Test settings.} In their original papers, \svdd and \gan are trained on what we refer to as Clean (inlier only) data, and tested on a disjoint test dataset.
In contrast, 
 \rda and \rand consider the \textit{transductive} setting where the train data is the same as the test data, containing inliers as well as outliers, which we refer to as Polluted.
It is often understood to be more challenging than the Clean setting for unsupervised OD.
For completeness, we evaluate these methods under both settings.

\begin{figure}[!t]
	\vspace{-0.1in}
	\centering
	\begin{tabular}{cc}
		\hspace{-0.2in}
		\includegraphics[width=0.45\textwidth, height=1.5in]{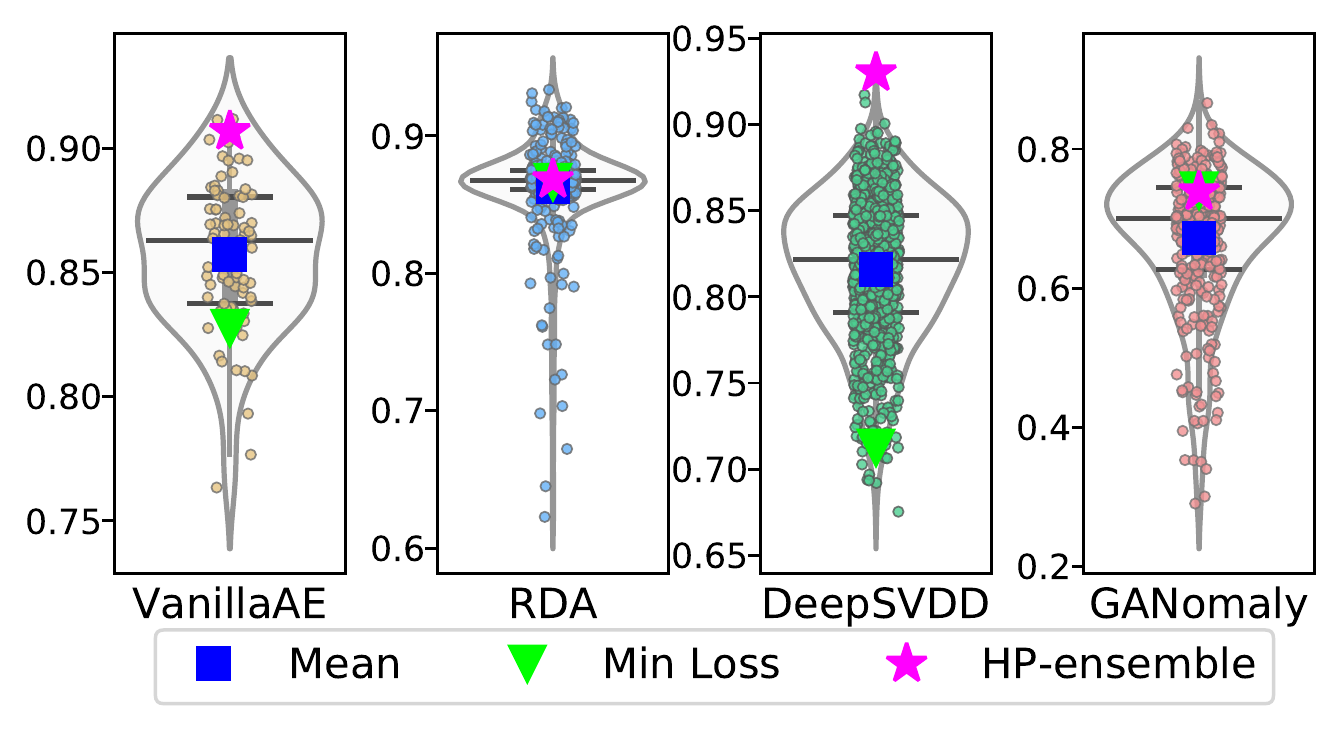} &
		\hspace{-0.1in}
		\includegraphics[width=0.55\textwidth, height=1.5 in]{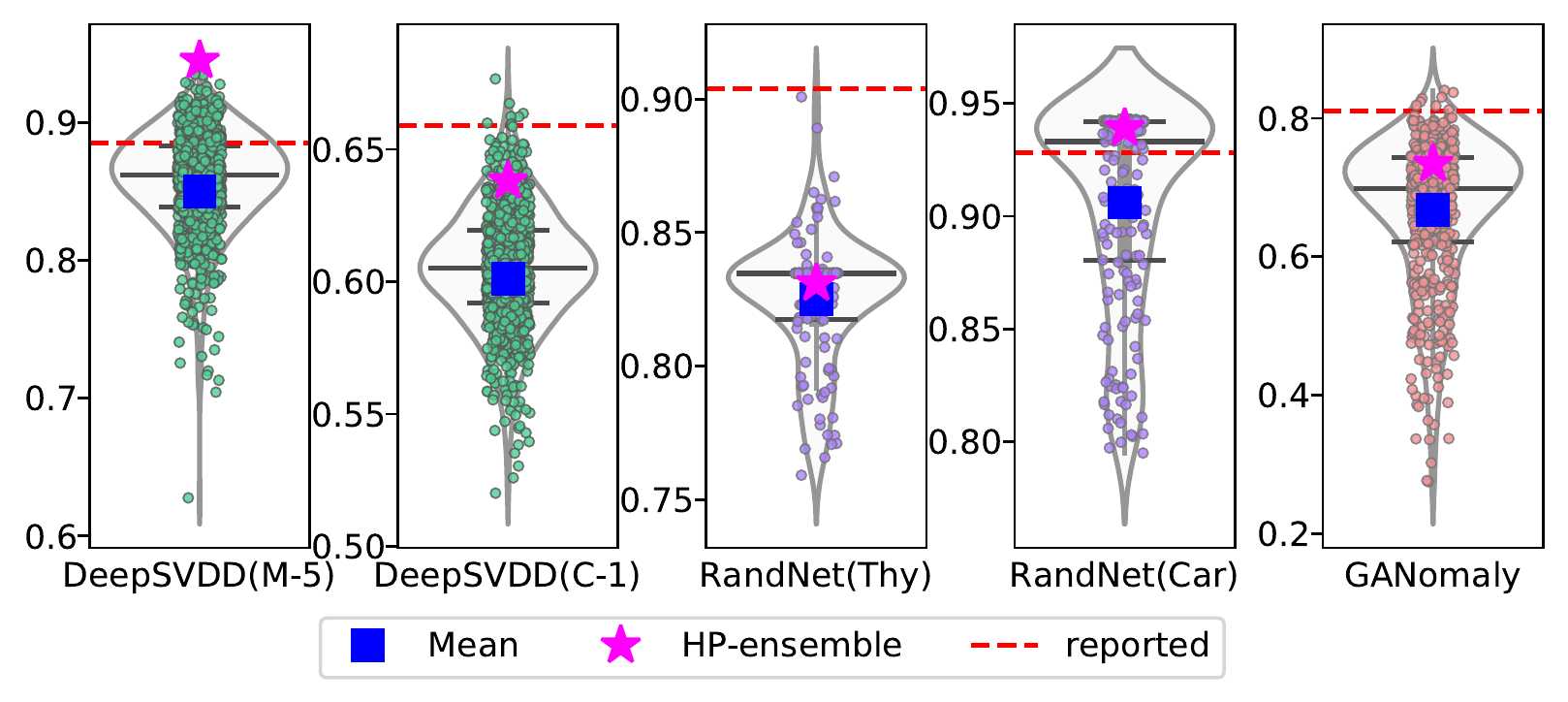} 
	\end{tabular}
	\vspace{-0.15in}
	\caption{\label{fig:mnist4inall} (left) AUROC performance of deep OD methods with different HP configurations (circles) on MNIST-4 showcase notable variation (i.e., sensitivity). (See footnote 2 below.) %
		(right) Similar results on additional datasets; 
		\svdd on MNIST-5 and CIFAR10-auto, \rand on Thyroid and Cardio, and \gan on MNIST-4out.
		Reported values (dashed lines) are overestimates of Mean ($\square$). Hyper-ensemble (\FiveStarOpen) improves notably over Mean. 
		}
	\vspace{-0.2in}
\end{figure}

{\bf Datasets.~} For evaluation we consider both image point datasets.
As in the original papers \cite{akcay2018ganomaly,conf/icml/RuffGDSVBMK18}, we use MNIST and CIFAR10 to construct OD tasks, 
except for \rand which uses fully-connected rather than convolutional layers and is
originally 
tested only on point-cloud benchmarks.

{\bf Metrics.~} 
Across all HP settings of a deep OD method, we report the mean AUROC, standard deviation (stdev), as well as minimum and maximum. Mean corresponds to the \textit{expected} performance when a HP config is randomly picked from our (multi-dimensional) grid. We also report the average mean and its stdev over 3 repeated runs. 
For datasets that are directly comparable to those in the original papers, we contrast the mean model performance that we obtain to that reported value for author-recommended HPs.
We compare to a simple model (i.e. HP) selection heuristic, which 
selects the (one) model with the lowest loss/objective value. 
In addition, instead of selecting one, we average the scores from \textit{all} configurations and report the AUROC of this hyper-ensemble.

Over five deep OD methods, tens to hundreds of HP configs, multiple initializations, Clean and Polluted settings, and various datasets, we have trained 
a total of more than {35,000} models. As such, our study constitutes the first large-scale HP sensitivity analysis in the deep outlier mining literature.

\subsection{Results and Observations}
\label{ssec:findings}

Fig. \ref{fig:mnist4inall}(left) provides the AUROC performances\footnote{\label{note2}We remark that although the methods are presented side-by-side, the goal here is \textbf{not} to compare them ``head-to-head'' but rather, to analyze each one's performance variability by varying HPs on individual datasets.
} 
for \vae, \rda, \svdd and \gan across all HP configurations (circles), for 
the Polluted 
setting (See Appx. \ref{ssec:hpa} Fig. \ref{fig:mnist4Clean} for Clean setting.) 
on MNIST-4 dataset where digit `4' images are designated as  inliers and the rest nine classes are down-sampled at 10\% as outliers, as in the \svdd paper \cite{conf/icml/RuffGDSVBMK18}. Horizontal bars mark the 1st, 2nd (i.e. median), and 3rd quartiles, square the mean AUROC across HPs, triangle the selected model by lowest loss, and star the AUROC of the hyper-ensemble. (Corresponding plots over 3 runs per method are in Appx. \ref{ssec:hpa} Fig. \ref{fig:mnist4in}.) 
Table \ref{tab:mnist4in} provides summary statistics as well as the hyper-ensemble performance for comparison for
 both the (left) Clean  and (right) 
 Polluted settings.


First, in Fig. \ref{fig:mnist4inall} we observe that all methods exhibit notable variability in performance across HPs, with many  worse-than-average configurations; e.g. for \gan, as well as many better-than-average models; e.g. for \rda.
Mean performance is considerably lower than the best model's, illustrating the opportunity or room for improvement.
As one would expect, the performances in Table \ref{tab:mnist4in} are lower in the Polluted setting as compared to Clean (less so for the robust \rda),  while sensitivity (i.e. stdev) is comparable or only slightly higher; showing that HP choice is critical 
under both settings.  

We also find that selecting a model by the value of loss ($\triangledown$), despite requiring to train and choose from all models,  often is much worse than the mean, i.e. random picking a (single) model ($\square$); proving this simple heuristic ineffective.
On the other hand, the hyper-ensemble (\FiveStarOpen) outperforms the mean in all cases, with quite small standard deviation across runs (i.e. low sensitivity to initialization).

The reported performance ($0.949\pm0.008$) of \svdd on MNIST-4 (under Clean) \cite{conf/icml/RuffGDSVBMK18} is similar 
yet somewhat optimistic over the mean value we obtain. MNIST-4 is an easy task for \svdd since mean AUROC is already around $0.924$. Thus, we set up MNIST-5 (with digit `5' as inliers) with reported AUROC 
$0.885\pm0.009$, as well as CIFAR10-auto (with class `automobile' as inliers) (in both cases rest of the classes are subsampled at 10\% each as outliers) with reported AUROC 
$0.659\pm0.021$.
We run \svdd on both datasets under all 864 configurations. As shown in Fig. \ref{fig:mnist4inall} (right),
our mean AUROCs are 
$0.857\pm0.037$ for MNIST-5, and
$0.605\pm 0.024$ for CIFAR10-auto. (See Appx. \ref{ssec:hpa} Fig. \ref{fig:mnist5inCIFARauto} for all 3 runs.)
The reported performances for the ``recommended'' HPs\footnote{{Besides {fixed} values for various HPs,  Ruff \textit{et al.} \cite{conf/icml/RuffGDSVBMK18} recommend HPs that \textit{differ by dataset}; e.g. on MNIST they use 2 CNN modules w/ size 8 and 4 filters, on CIFAR10 they use 3 modules w/ size 32, 64, and 128 filters. 
			}} in \svdd (dashed red lines) appear to be optimistic over the mean, i.e. what one would expect by random choice (in the absence of any labels or other strategies). 

\begin{table}[!t]
	\vspace{-0.1in}
	\setlength{\tabcolsep}{5pt}
	\caption{
		Basic stats of AUROC ($\%$) distribution over varying HPs  
		on MNIST-4  
		under  (left) Clean  and (right) Polluted  settings.
		There is significant gap between the best and worst HP settings, with notable stdev around the mean. 
		Polluted results have lower mean, and comparable or slightly higher variance.
		Hyper-ensemble outperforms random choice (i.e. mean), with low variability by initialization.
		\label{tab:mnist4in}
	}
	\vspace{-0.05in}
	\hspace{-0.05in}
	\centering
	{\scalebox{0.64}{
			\begin{tabular}{l|llr|c|c||llr|c|c}
				\toprule
				\textbf{Method}     & \multicolumn{2}{c}{\textbf{Min\&Max}}  & \textbf{Mean\&Std.}                       & \begin{tabular}[c]{@{}c@{}}\textbf{Mean}\\ (avg. 3 runs)\end{tabular} & \begin{tabular}[c]{@{}c@{}}\textbf{Hyper-ens. Mean}\\   (avg. 3 runs)\end{tabular} & \multicolumn{2}{c}{\textbf{Min\&Max}}  & \textbf{Mean\&Std.}                       & \begin{tabular}[c]{@{}c@{}}\textbf{Mean}\\ (avg. 3 runs)\end{tabular} & \begin{tabular}[c]{@{}c@{}}\textbf{Hyper-ens. Mean}\\   (avg. 3 runs)\end{tabular} \\
				\midrule
				Vanilla AE & 87.41        & 98.19       &  93.12$\pm$2.91                                                                  & \textbf{93.12}$\pm$0.008                                              & \textbf{95.30}$\pm$0.03   & 76.34       & 91.20       &  85.73$\pm$2.95 & \textbf{85.76}$\pm$0.08                                               & \textbf{90.46}$\pm$0.12                                                           \\
				RDA        & 81.79        & 95.96       &  88.94$\pm$2.43                                                                  & \textbf{88.95}$\pm$0.02                                               & \textbf{89.39}$\pm$0.02   & 62.29        & 93.35       &  86.30$\pm$3.68 & \textbf{86.24}$\pm$0.06                                               & \textbf{86.82}$\pm$0.04                                                            \\
				DeepSVDD   & 75.75        & 96.58       &  92.39$\pm$2.02 & \textbf{92.37}$\pm$0.02                                               & \textbf{95.94}$\pm$0.05   & 67.54        & 91.71       &  81.65$\pm$4.15 & \textbf{81.66}$\pm$0.10                                               & \textbf{93.07}$\pm$0.70                                                          \\
				GANomaly   & 17.49        & 95.73       &  78.90$\pm$16.63                                                                  & \textbf{78.26}$\pm$0.51                                               & \textbf{86.87}$\pm$0.31  & 29.06        & 86.60       & 67.20$\pm$10.30 & \textbf{67.45}$\pm$0.32                                               & \textbf{73.87}$\pm$0.52                                                           \\
				\bottomrule                                                
			\end{tabular}
	}}
	\vspace{-0.15in}
\end{table}



The optimistic reporting trend holds for \gan and \rand as well.
Following their original paper \cite{akcay2018ganomaly}, and different from \cite{conf/icml/RuffGDSVBMK18}, we set up MNIST-4out dataset to contain digit `4' images this time as the outliers and the rest nine classes as inliers.
\gan performances across 162 HP configs are shown in Fig. \ref{fig:mnist4inall} (right), where the reported result (red line) in \cite{akcay2018ganomaly} of AUROC $0.795$ appears notably higher than the mean $0.668\pm0.105$ that we obtain. (See Appx. \ref{ssec:hpa} Fig. \ref{fig:ganomaly} for all 3 runs.)

For \rand, we first find and verify a third-party implementation, as the authors could not publicly share theirs, by replicating similar performances to those reported in \cite{conf/sdm/ChenSAT17} using the author-recommended HP settings on all 8 datasets.  Then for Cardio and Thyroid datasets, we train 192 \rand ensembles with varying HPs under Polluted  as in the original paper.
Results from one run are shown in Fig. \ref{fig:mnist4inall} (right). (See Appx. \ref{ssec:hpa} Fig. \ref{fig:randnet} for all 3 runs.) 
On Cardio, AUROC mean is $0.894\pm0.045$ vs. $0.929$ (reported), and
Thyroid mean is $0.822\pm0.026$ vs. $0.904$ (reported).




In summary, the take-aways from our analysis are as follows.
First, it is clear that deep OD models are sensitive to their HP settings, showing that model/HP selection is inevitable. The mean/expected value (of random choice) can be quite away from the best model, motivating this line of research. Second, the recommended settings in recent deep OD papers are arguably optimistic; otherwise more transparency into their selection mechanism is warranted.
Finally, hyper-ensemble performance is superior to the mean, possibly owing to different HPs implicitly imposing diversity among constituent models which helps improve detection. Besides performance improvement, hyper-ensembling
obviates model selection by joining all models rather than choosing one and is not much sensitive to initialization -- setting the stage for our proposed HP-robust OD method \method.

\section{\method: A Deep Hyper-ensemble for Hyperparameter-Robust OD}
\label{sec:dod}

\subsection{Motivation and Overview}
\label{ssec:overview}
\vspace{-0.05in}

The main research question we consider is \textbf{RQ1) how to design an unsupervised deep OD model that is robust to its HPs}, i.e. an OD method that has stable, low-variance predictions under varying HPs.
Motivated by our sensitivity analysis in Sec. \ref{sec:sensitive}, we propose a deep autoencoder (AE) hyper-ensemble model that combines scores from AE models with different HP configurations.

\begin{definition}[Hyper-ensemble]
Given a model family $\mathcal{M}$ with $H$ HPs, let $\blambda \in \realR^H$ denote a specific setting of the HPs.
A hyper-ensemble averages the output (outlier scores) from a finite number of base models with $m$ different config.s, i.e. $\frac{1}{m} \sum_{i=1}^m \mathcal{M}(\mathbf{x}; \blambda_i)$ for point $\mathbf{x}$.	
\end{definition}

Hyper-ensembles exhibit multiple advantages. First, and to our end goal, they ease the model/HP selection burden, bypassing the ``choice paralysis''. They are less sensitive to random initializations, with lower variance in performance. Moreover, they can even boost detection performance thanks to the diversity offered by different HPs and ensemble prediction. 

The caveat is that deep ensembles are computationally expensive to train.
As such, the second research task we tackle is 
\textbf{RQ2) how to 
speed up hyper-ensemble training}.
To that end, we propose novel architectural and training strategies. 
In a nutshell, these strategies involve three main ideas: 
(1) we design a multi-layer AE architecture with \textit{skip} connections, denoted \aes, that helps hyper-ensemble, under a \textit{single} model, \textit{varying-depth} AEs with \textit{shared} parameters;
(2) we employ batch ensemble training \cite{conf/iclr/WenTB20} of multiple \aes models using \textit{varying size zero-masking} that helps hyper-ensemble \textit{varying-width} AEs all trained \textit{simultaneously} with \textit{shared} parameters; and
(3) we train each \aes on a \textit{subsample} and use out-of-sample scoring.
In effect, these strategies allow us to build fewer models, with fewer total number of parameters, on less training data---{taking only $2$-$10$\%}
of the time that the na\"ive ensemble training would take
 where each model is trained independently.

\begin{figure}[!t]
	\vspace{-0.1in}
	\centering
	\includegraphics[width=\linewidth]{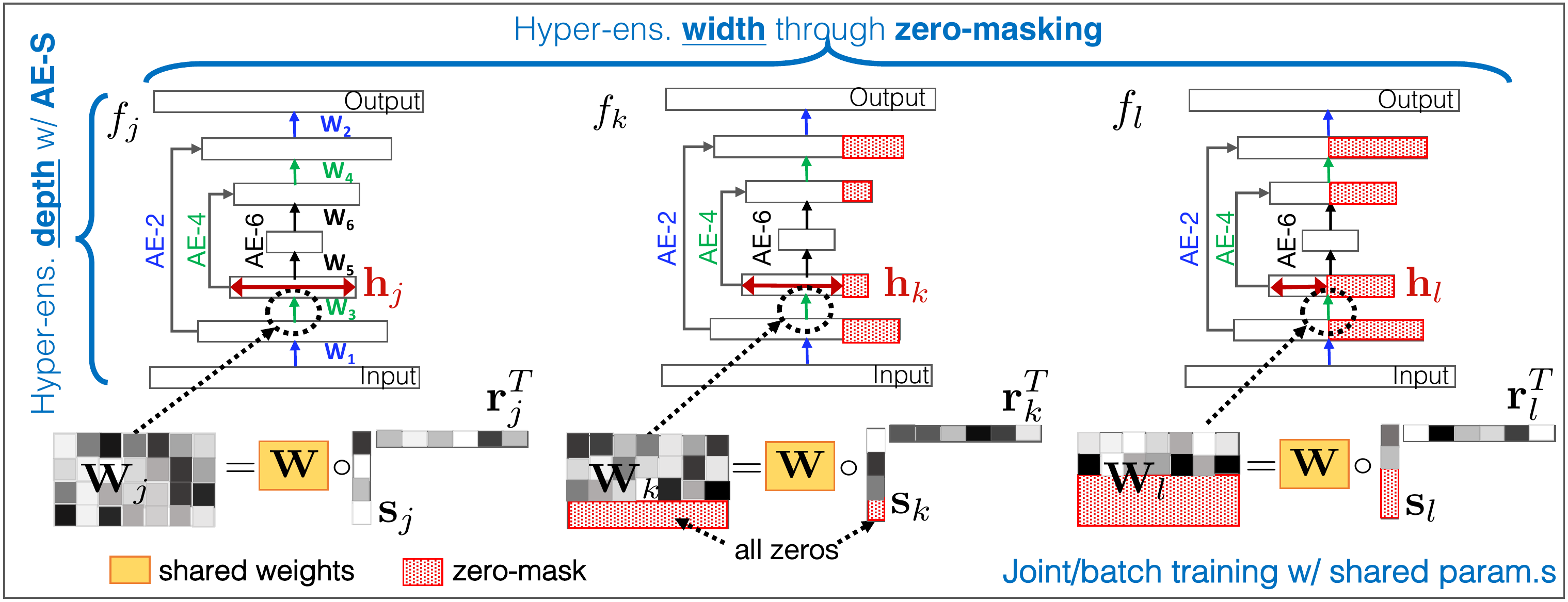}
	\vspace{-0.25in}
	\caption{Main design elements in \method:  \textbf{(1) AE with Skip links (AE-S)}: Each \aes, denoted $f$, hyper-ensembles multiple AE models with various depths (in the figure, AE-2, 4 and 6), with \textit{shared parameters}; e.g. for $f_j$, layer 2 weights $\bm{W}_j$ (in dashed circle) is shared among \underline{different-depth} AEs (i.e. AE-4 and AE-6).
		 \textbf{(2) Batch ensemble (BE) training with zero-masks}: Multiple \aes models with \textit{shared parameters} are trained simultaneously, having various widths thanks to zero-masking; e.g. at layer 2, $\bm{W}$ is shared by all three \underline{different-width} \aes models in the batch (denoted $f_j$, $f_k$, $f_l$), where hidden sizes follow $|\bm{h}_j|>|\bm{h}_k|>|\bm{h}_l|$ through varying size zero-masks on $\bm{s}_k$ and $\bm{s}_l$. ($\bm{s} \bm{r}^T$ depicts an outer product, and $\circ$ denotes Hadamard/element-wise matrix multiplication.) 
		}
	\vspace{-0.2in}
	 \label{fig:chart}
\end{figure}

Fig. \ref{fig:chart} illustrates the 
main design elements
of \method pictorially. We present details of the architecture and the 
training as follows.

\vspace{-0.05in}
\subsection{Design Strategies for Speeding up Ensemble Training}
\label{ssec:design}
\vspace{-0.05in}

\subsubsection{Hyper-ensembling depth: One model for multiple depths}
\label{sssec:dep}
\vspace{-0.05in}

Skip or shortcut connections are applied in NNs 
for various purposes; e.g., in 
ResNet \cite{conf/cvpr/HeZRS16} for solving the depth degradation problem, in JK-Net \cite{Xu2018vn} for representation learning with varying localities, in U-Net \cite{unet} and stochastic depth NNs \cite{conf/eccv/HuangSLSW16} 
for training models with adaptive depth, and so on.

Due to the ``hourglass'' structure of the AE, skipping one layer in an AE would cause dimensionality mismatch for the next layer. Instead, we create long shortcuts by skipping the middle layers while keeping the outer encoder-decoder pairs symmetrically, and refer to this architecture as \aes.  

 An illustration of hyper-ensembling depth can be found in Fig. \ref{fig:chart}; see e.g. the \aes denoted by $f_j$. Given a $2L$-layer \aes ($L$ encoder and $L$ decoder layers), there are $2L$ trainable weight matrices. For input $\mathcal{D}$, we generate $L$ outputs by allowing the input to pass through the outermost encoder-decoder pair (which we denote as $AE\text{-}2$), the two outermost encoder-decoder pairs ($AE\text{-}4$), and so on, until the full structure ($AE\text{-}2L$) is traversed by the input.
 In effect, this trains $L$ different AEs under \textit{one} \aes model.
  Then, the overall loss $\mathcal{L}_{\text{\aes}}$ of the depth-hyper-ensembling \aes is the summation of 
  the member AEs'
  reconstruction errors (denoted $\Delta Err$). Eq. \eqref{eq:AEs_loss} gives an example of the loss function for a $6$-layer \aes as shown in Fig. \ref{fig:chart},
\begin{equation}
 \scalemath{0.975}{
	\label{eq:AEs_loss}	
	\mathcal{L}_{\text{\aes}}
	=  \Delta Err_{AE\text{-}2}(\mathcal{D}; [\mathbf{W}_1\text{:}\mathbf{W}_2]) 
	+
	 \Delta Err_{AE\text{-}4}(\mathcal{D}; [\mathbf{W}_1\text{:}\mathbf{W}_4]) 
	 +
	 \Delta Err_{AE\text{-}6}(\mathcal{D}; [\mathbf{W}_1\text{:}\mathbf{W}_6]) 
}
\end{equation}
\aes is an implicit hyper-ensemble which allows simultaneous/joint training of AEs with various depths. It is computationally efficient thanks to parameter sharing, as the outer layers' weights are reused and tuned among different members. 
Each ensemble member can also play a regularization effect 
and prevent an AE from producing low scores for outliers due to overfitting. (See Appx. \ref{ssec:reg}). 

\subsubsection{ Hyper-ensembling width: Zero-masked joint training}
\label{sssec:wid}
\vspace{-0.05in}

BatchEnsemble \cite{conf/iclr/WenTB20} (BE) is a state-of-the-art parameter-efficient deep ensemble training approach. However, 
off-the-shelf, it does not allow for {\em hyper-}ensebling, that is, all models in the batch are trained with the same HP configuration, and share equal size parameters.
 Our approach builds on BE, adapting it to simultaneously train varying-width \aes with shared parameters.

First, we briefly review the BE architecture with $K$ AEs. For a specific layer in BE, weight $\bw \in \realR^{m\times r}$ is shared across all $K$ individual members, while each member $i$ maintains two trainable rank-1 vectors: $\bss_i \in \realR^m$ and $\br_i \in \realR^r$. The outer product ($\cdot$) of $ \br_i$ and  $\bss_i^T$ creates a ``mask'' (a matrix in $\realR^{m\times r}$) onto the shared weight $\bw$ and generates individual weight $\overline{\bw_i}$. For the input mini-batch  $\mathbf{X}_i \in \realR^{n \times m}$, the forward propagation computation is 
\begin{equation}
	\Phi \big(\mathbf{X}_i \overline{\bw_i} \;  \big) = \Phi \big( \mathbf{X}_i \; (\bw \circ ( \bss_i \cdot \br_i^T ))  \big) = \Phi \big( (\mathbf{X}_i \circ \bss_i) \bw \circ \br_i \big) \;,
	\label{eq:be}
\end{equation}
where $\Phi$ is the activation function, $\circ$ is the element-wise product, and $\br_i$ and $\bss_i$ are broadcasted row-wise or column-wise depending on the shapes at play. To vectorize the calculation for all $K$ members in the ensemble, two matrices $\bSS \in \realR^{K \times m}$ and $\bR \in \realR^{K \times r}$ are constructed with rows of $\bss_i$ and $\br_i$, respectively. Thus, for the input 
$\mathbf{X} \in \realR ^{Km \times n}$, the BE layer computes the next layer as
\begin{equation}
	\Phi \big( (\mathbf{X} \circ {\bSS}) \bw \circ \bR \big) \;.
\end{equation}
Since $\mathbf{X}$ is composed of $K$ mini-batching $\mathbf{X}_i$'s tiling up, each member can utilize a different mini-batch of input. The $K$ members are training simultaneously in parallel using one forward pass, thus relieving the memory and computational burden of traditional ensemble methods.

While BE itself is not a hyper-ensemble, we can leverage this architecture to aggregate networks of different widths, also in an efficient way. The widths of a neural network layer correspond to the rows in the weight matrix. For a specific layer, instead of directly operating on $\bw$, we instead initialize a zero-one masking vector $\mathbf{\balpha}_i \in \{0,1\}^r$. $\mathbf{\balpha}_i$ is to element-wise multiply with the $\br_i \in \realR^r$, such that the operation $\bw \circ \bss_i \cdot (\br_i^T \circ \mathbf{\balpha}_i^T)$ will create a masking matrix of size $(m \times r)$ and sparsify the individual weight $\overline{\bw_i}$ by allowing zero'ed-out rows. In vectorized notation, the matrix $\mathbf{A} \in \{0,1 \}^{K \times r}$ is composed of $K$ distinct masking vectors. Then, the zero-masked BE forward propagation becomes
\begin{equation}
	\Phi \big( \mathbf{X} \circ \bSS ) \bw \circ (\bR \circ \mathbf{A}) \big)\;.
\end{equation}
Sparsifying neural networks has been shown to decrease storage and improve training efficiency, 
with many algorithms built to wisely prune the neural network \cite{hoefler2021sparsity}. While zero-masked BE is similar to creating neural networks of varying density, our main goal is to obtain models with different widths (i.e. hidden sizes), such that when trained under BE, creates a width-hyper-ensemble.
Specifically, we construct the zero-masked BE layer with the maximum width and specify the zeros in respective $\mathbf{\balpha}$ vectors,  such that the masked-out individual weights correspond to varying-width models. 

The zero-masked BE is 
efficient as $\mathbf{A}$ is fixed throughout training. The element-wise product of the masking incurs little extra time during forward and backward propagation, while all the (rank-1) vectors are cheap to store compared to 
separate weight matrices as in traditional ensemble training.

\subsubsection{\method: The overall hyper-ensemble}

Let $H$ denote 
 the total number of HPs for an \aes, where
we define a grid of values for each HP; e.g. \dep$=$$[4,6,8]$, \wid$=$$[64,128,256]$, 
\lr$=$$[1e^{-3}, 5e^{-4}]$, \dr$=$$[0.0, 0.2]$, etc.

For the two HPs, \dep and \wid, we specify the \aes with the largest depth value (following Sec. \ref{sssec:dep}, say $2L$) in the grid and leverage the skip connections to obtain the smaller-depth AEs.
Similarly, we specify each \aes in the zero-masked BE with the largest width value (following Sec. \ref{sssec:wid}, say $K$) in the grid and leverage the zero-masking to obtain other, smaller-width AEs.

Then, a zero-masked BE trains in parallel $K$ varying-width \aes models, each being an ensemble of $L$ varying-depth AEs. 
Outlier score for a point $\mathbf{x}$ is averaged across all  $KL$ AE models as
\begin{equation}
s(\mathbf{x}) = \frac{1}{KL}\sum_{i=1}^{K}{s_i(\mathbf{x})}\;, \;\; \text{where} \; s_i(\mathbf{x}) = \sum_{d=1}^L \lVert  \mathbf{x} - f_i^{(AE\text{-}2d)}(\mathbf{x}; {\bm{\lambda}})  \rVert^2 \;, 
\end{equation}
where $f_i$ is the $i$'th \aes member, $f_i^{(AE\text{-}2d)}$ denotes the AE associated with depth $2d$ within $f_i$, and 
$\bm{\lambda}$ is a vector depicting a specific configuration of all the remaining $(H$$-$$2)$ HPs; e.g. $[$\lr$=$$1e^{-3}$, \dr$=$$0.2$, etc.$]$. 
Denoting the total number such configurations by $B$,
\method averages scores, i.e. the $s(\mathbf{x})$ values, from $B$ different zero-masked BEs as the final outlier score of $\mathbf{x}$.



\subsubsection{Further speed up by subsampling}
As shown in Eq. \ref{eq:be}, BE allows mini-batching, where each data point can be used by one or several different ensemble members. This makes BE a natural fit to subsampling, which further expedites the training procedure. To this end, 
we create $\{\mathbf{X}_i^{in}, \mathbf{X}_i^{out} \}_{i=1}^K$ splits of the training data, where
for each \aes member $i$, we divide the training data into $\mathbf{X}_i^{in}$ and $\mathbf{X}_i^{out} = \mathbf{X}_i / \mathbf{X}_i^{in}$ and solely train on mini-batches from $\mathbf{X}_i^{in}$. 
We then compute the out-of-sample\footnote{Applicable to transductive OD only; for inductive OD, a point is scored by all ensemble members.}
outlier score of point $\mathbf{x}$ as
$s(\mathbf{x}) = \frac{1}{K'L} \sum_{i=1}^{K} \identity (\mathbf{x} \in \mathbf{X}_i^{out}) s_i(\mathbf{x})$, 
	 where $K' = \sum_{i=1}^{K} \identity (\mathbf{x} \in \mathbf{X}_i^{out})$.


\section{Experiments}
\label{sec:experiments}
\subsection{Experimental Setup}
\label{expsec:setup}

{\bf Baselines.~} 
We compare to SOTA \textit{deep} OD methods \vae, \rda \cite{conf/kdd/ZhouP17}, \svdd \cite{conf/icml/RuffGDSVBMK18} and \rand \cite{conf/sdm/ChenSAT17}, the last of which is a deep AE ensemble. We also include the tree-ensemble Isolation Forest \cite{liu2008isolation} (\ifor) which stands as the SOTA among \textit{traditional} detectors \cite{emmott2015meta}.
Besides \method without subsampling, we 
experiment with two subsampling versions, denoted \method-$\delta$, for sampling rates $\delta \in \{0.1,0.5\}$. We also compare to the na\"ive \method with independent training, denoted \imethod.

{\bf Configurations.~} The baselines exhibit 2-8 HPs, per which we define a small grid of values (See Appx. \ref{ssec:setdetails} Table \ref{tab:HP overview}).
We report the expected AUROC performance, i.e., averaged across all configurations in the grid, along with the standard deviation. 
For \method (and variants) we set $L$$=$$6$, and $K$$=$$8$; the other HP config.s are listed in Appx. Table \ref{tab:HPROBODoverview}.
To measure sensitivity to random  initialization, we average performance across 3 runs.  
All models are trained on a NVIDIA RTX A6000 GPUs server.

\begin{wraptable}{r}{4.25cm}
	\setlength{\tabcolsep}{1.2pt}
	\vspace{-0.25in}
\centering
\caption{Dataset statistics. }
\vspace{0.05in}
	\centering
	\hspace{-0.1in}
	{\scalebox{0.8}{
    \begin{tabular}{lrrrr}
    \toprule
    \textbf{Name} &  $\#$ \textbf{pts.} &  \textbf{dim.} &   \textbf{outl.}$\%$ \\
    \midrule
    MNIST-4 & 6426 &  1$\times$28$\times$28 & 10.0 \\
    MNIST-5 & 6426 &  1$\times$28$\times$28 & 10.0 \\
    MNIST-8 & 6426 &  1$\times$28$\times$28 & 10.0 \\
    CIFAR10-0 & 5500 & 3$\times$32$\times$32 & 10.0  \\
    CIFAR10-1 & 5500 & 3$\times$32$\times$32 & 10.0  \\
    Thyroid & 3772 & 6 & 2.5 \\
    Cardio & 1831 & 21 & 9.6 \\
    Lympho & 148 & 18 & 4.1 \\
    \bottomrule
    \end{tabular}
}}
	\vspace{-0.25in}
    \label{tab: datasetoverview}
\end{wraptable} \textbf{Datasets.} We conduct experiments on 5 image datasets from MNIST and CIFAR10, as well as 3 tabular datasets from the ODDS repository.\footnote{\url{http://odds.cs.stonybrook.edu/}} MNIST and CIFAR10 are multi-class, where we pick one class as the inliers and subsample the rest at 10\% each to constitute outliers.
Datasets have varying outlier \% and size, with images having high dimensionality. (See Table \ref{tab: datasetoverview}; MNIST-4, -5, -8 depict respective digits as inliers. CIFAR10-0 and -1 refer to \air and \auto class as inliers, respectively.)
Details on dataset description and preparation can be found in Appx.
\ref{sssec:datadetails}. The experiments are all conducted under the Polluted (i.e., transductive) setting.

\subsection{Results}
\label{expsec:results}

With main results shown in Table \ref{tab:AUC result}, we want to answer the following questions: \textbf{Q1. How does \method compare to state-of-the-art (SOTA) OD methods? }
\method achieves superior performance to all deep OD baselines on MNIST and tabular datasets, and competitive performance on CIFAR10 datasets against the overall runner-up \rand.
Notably, \method performs similarly or even better than \imethod, the latter 
potentially owing to the guarding effect of parameter sharing  against overfitting.
Moreover, similar performance can be retained under subsampling, where we train each hyper-ensemble member with 50\% or even 10\% of the data.

\textbf{Q2. How much does \method's performance vary by initialization?} The standard deviation (stdev) of the deep OD baselines is notably large; it is slightly smaller for the ensemble model \rand, whereas \method has significantly smaller stdev, with near-zero sensitivity to random initialization.
The sensitivity of deep baselines suggests that with an arbitrary choice of HPs in the absence of any other guidance, one may acquire much less satisfactory outcomes than the average AUROC. 

%

An interesting observation is regarding the traditional \ifor baseline, which excels on (low-dim.) tabular datasets, and remains competitive on MNIST-4 and CIFAR10-auto. It also shows relatively small variance w.r.t. its (two) HPs. However, it is significantly inferior on other high-dim. image OD tasks, which may be attributed to its lack of representation learning. Nevertheless, the competitiveness of this simple  baseline suggests that  traditional OD methods cannot be ignored in the `horse-race' of developing new deep OD methods, not only for their competitiveness but also for their robustness to only-a-few HPs that makes them easy to employ by practitioners.

\begin{table*}[!t]
	\vspace{-0.1in}
	\centering
	\caption{AUROC ($\%$) performance of OD methods. 
		Baselines (top) avg.'ed across HP config.s, \method and variants (bottom) avg.'ed over 3 runs w/ random init.s.; $\pm$ one stdev. Entries in {\color{green} green} depict the method w/
		least variance.
		Highlighted in bold and underline are the \textbf{best} and \underline{runner-up}. }
	\resizebox{\columnwidth}{!}{
	\begin{tabular}{lrrr|rr|rrr}
		\toprule &  MNIST-4 & MNIST-5 & MNIST-8 & CIFAR10-air & CIFAR10-auto & Cardio & Thyroid & Lympho \\
		\midrule
		\vae  & 81.4$\pm$9.4& 73.6$\pm$10.1 & 83.3$\pm$4.6 & 
		\underline{60.3$\pm$2.0} &
		59.2$\pm$4.7 &
		87.1$\pm$7.7 &
		81.1 $\pm$8.5 &
		89.4$\pm$11.7 \\
		RDA & 79.7$\pm$11.2 &
		68.9$\pm$11.4   &
		82.6$\pm$10.2     & 
		53.9$\pm$8.8 &
		54.1$\pm$8.9 &
		78.4$\pm$12.1 &
		80.9$\pm$5.3 &
		78.0$\pm$12.2 \\
		\svdd &81.9$\pm$4.3 &
		75.7$\pm$3.8   &
		85.3$\pm$3.9 & 
		55.6$\pm$3.1 &
		59.0$\pm$3.0 &
		54.6$\pm$7.9 &
		67.5$\pm$15.1 &
		66.5$\pm$15.6 \\
		\rand  & 
		85.3$\pm$3.1   &
		79.3$\pm$3.8    &
		85.4$\pm$2.2     &
		59.1$\pm$2.6  &
		59.5$\pm$3.8 &
		89.2$\pm$5.3&
		81.5$\pm$5.8 &
		92.3$\pm$8.1 
		\\
			\ifor & 84.2$\pm$0.8 & 70.1$\pm$1.5 & 70.5$\pm$1.2 & 42.9$\pm$0.5 &
			\textbf{62.7$\pm$0.8} &
			\textbf{94.1$\pm$0.9}&
			\textbf{97.9$\pm$0.4} &
			\textbf{99.5$\pm$0.3} \\
		\midrule
	
		\method & \cellcolor{green!30}\textbf{88.0$\pm$0.0} &
		\cellcolor{green!30}\textbf{81.4$\pm$0.1} &
	    \cellcolor{green!30}87.8$\pm$0.0 &
		\cellcolor{green!30}59.4$\pm$0.0 &
		\cellcolor{green!30}59.2$\pm$0.1 &
		\cellcolor{green!30}\underline{93.5$\pm$0.1} &
		86.1$\pm$0.6 &
		98.7$\pm$0.1 
		\\
		\method-0.5 & \underline{86.7$\pm$0.1} &
		\cellcolor{green!30}\underline{78.5$\pm$0.1} &
		\underline{88.1$\pm$0.1} &
		59.6$\pm$0.5 &
		58.4$\pm$0.3 &
		92.1$\pm$0.3 &
		87.9$\pm$1.3 &
		98.7$\pm$0.1 
		\\
		\method-0.1 & \cellcolor{green!30} \underline{86.7$\pm$0.0} &
		\cellcolor{green!30}\underline{78.5$\pm$0.1} &
		\textbf{88.2$\pm$0.1} &
		\cellcolor{green!30}59.3$\pm$0.0 &
		58.3$\pm$0.2 &
		91.8$\pm$0.3 &
		88.4$\pm$0.5 &
		\underline{99.0$\pm$0.1} 
		\\
			\imethod  &\cellcolor{green!30}84.5$\pm$0.0 &
			\cellcolor{green!30}74.6$\pm$0.1 &
			\cellcolor{green!30}87.0$\pm$0.0 &
			\cellcolor{green!30}\textbf{62.5$\pm$0.0} &
			\cellcolor{green!30}\underline{61.7$\pm$0.1}   &
			\cellcolor{green!30}87.1$\pm$0.1 &
			\cellcolor{green!30}\underline{93.0$\pm$0.0} &
			\cellcolor{green!30}98.9$\pm$0.0 
			\\
		\bottomrule
	\end{tabular}
	}
	\vspace{-0.15in}
	\label{tab:AUC result}
\end{table*}

%

\begin{wrapfigure}{l}{0.565\textwidth}
		\vspace{-0.15in}
	\centering
	\begin{tabular}{cc}
		\hspace{-0.1in}
		\includegraphics[width=0.283\textwidth]{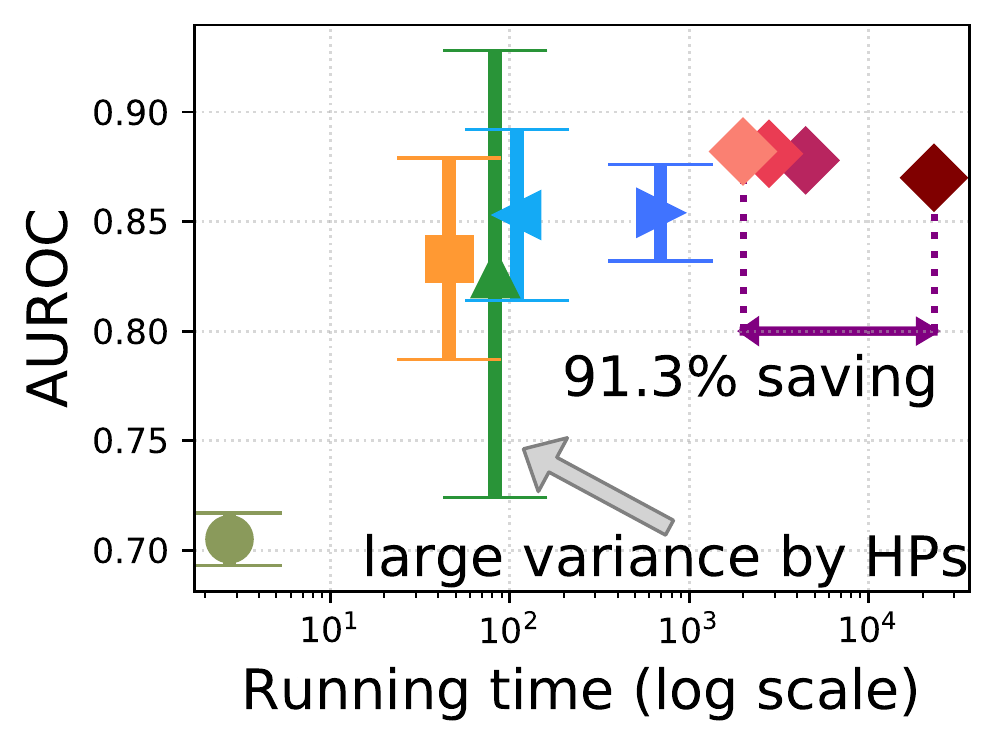} &
		\hspace{-0.25in}
		\includegraphics[width=0.283\textwidth]{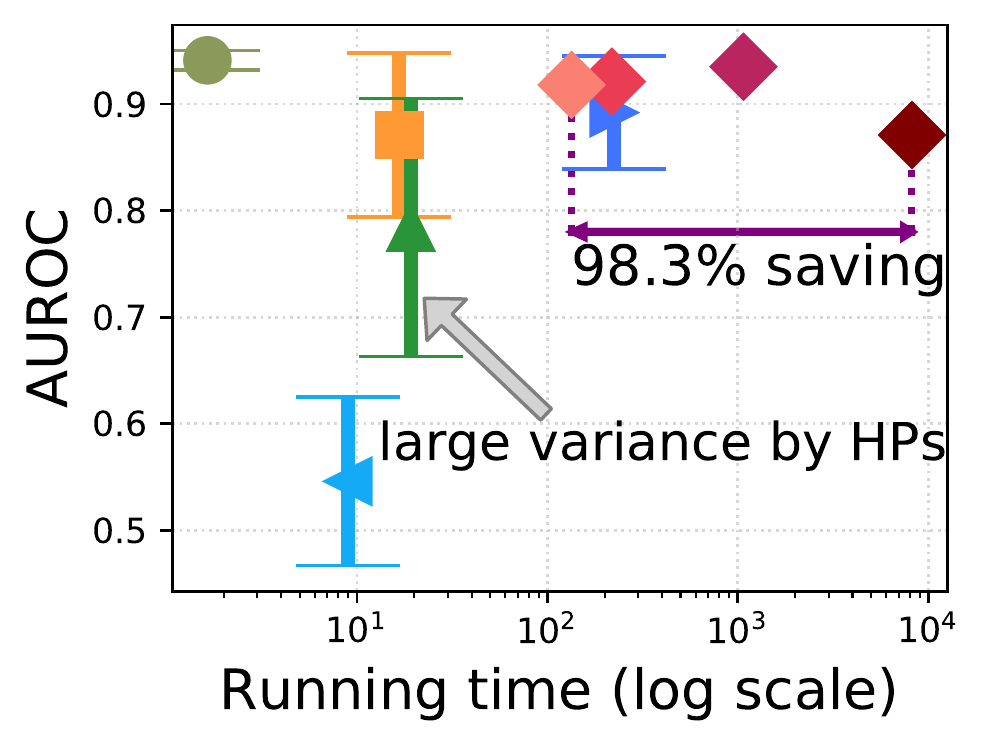} \\
	\multicolumn{2}{c}{	\hspace{-0.15in}\includegraphics[width=0.595\textwidth]{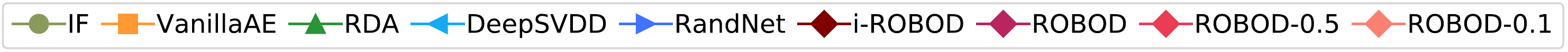} }
	\end{tabular}
	\vspace{-0.15in}
	\caption{\label{fig:scatter} Running time (in log scale) vs. AUROC performance of OD methods (symbols) on (left) MNIST-8 and (right) Cardio. Vertical bars depict one (1) stdev across HP config.s for the baselines.}
	\vspace{-0.1in}
\end{wrapfigure}

In Fig. \ref{fig:scatter} we show the running time (in sec.s) versus the performance for all OD methods (colored symbols) on one image and one tabular dataset for brevity. (See Appx. \ref{ssec:extra} Fig. \ref{fig:runtimevsauc} for remaining datasets.)
\method achieves higher performance, with lower, near-zero variability compared to deep baselines. Traditional \ifor, based on randomized trees, is very fast, yet it often underperforms on high dimensional tasks such as with images.

\textbf{Q3. What are the savings in running time compared to na\"ive hyper-ensemble w/ independent training? } We study the savings that \method offers compared to the na\"ive hyper-ensembling. As shown in Table \ref{tab:runtime}, it runs 3-10$\times$ faster across datasets. With subsampling at 10\%, \method-0.1 runs in about $1.6$-$10$\% of what it takes by \imethod.
Fig. \ref{fig:scale} (on MNIST-5) shows that while \imethod time increases by the number of models, i.e. larger $L$ and $K$ (fixing other HPs), \method takes near-constant time thanks to the batch/simultaneous training of 
varying-depth and -width members.


\begin{table}[!ht]
	\vspace{-0.15in}
	 \begin{minipage}[b]{0.6\textwidth}
	  \centering		
   \caption{Running time (sec.) of the na\"ive ensemble \imethod vs. \method. \method-0.1 that trains with 10\% subsampling offers 90-98\% relative savings in training time. }
    \vspace{-0.01in}
	\centering
	{\scalebox{0.623}{
   \begin{tabular}{l|rrrrrrrr}
   		\toprule &  MT-4 & MT-5 & MT-8 & CIF-air & CIF-auto & Cardio & Thy. & Lym. \\
	\midrule
                            \imethod    & 24192   & 24656   & 23165   & 38830   & 39562   & 8205    & 14313   & 1804   \\\hline
                            \method     & 4521    & 4630    & 4456    & 10721   & 10722   & 1070    & 2049    & 180    \\\hline
                            \method-0.1 & 2241    & 2141    & 1997    & 2834    & 3003    & 134     & 295     & 44     \\
                            Savings (\%)                    & \cellcolor{gray!30}90.74& \cellcolor{gray!30}91.32 & \cellcolor{gray!30}91.38 & \cellcolor{gray!30}92.70 & \cellcolor{gray!30}92.41 & \cellcolor{gray!30}98.37 & \cellcolor{gray!30}97.94 & \cellcolor{gray!30}97.56 \\ \bottomrule
   \end{tabular}
}}
    \label{tab:runtime}
      \end{minipage} \hfill
	  \begin{minipage}[t]{0.4\textwidth}
 	  \centering
	  \includegraphics[width=1.0\textwidth]{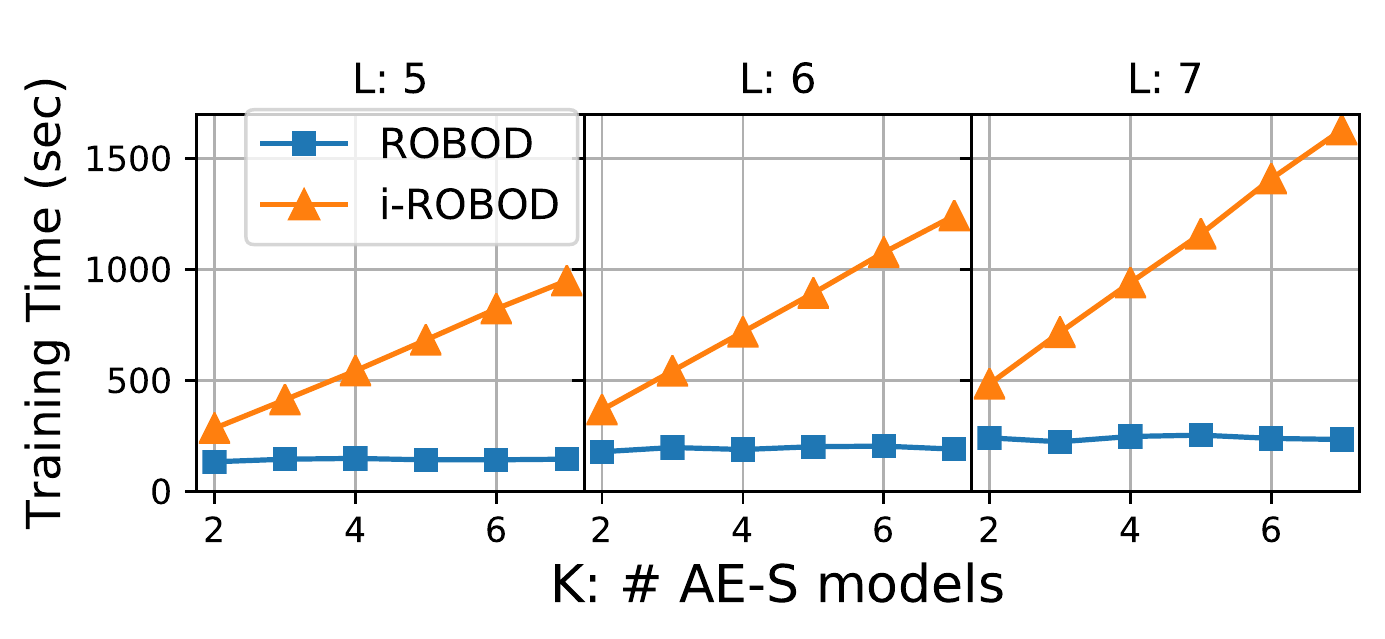}
	  \vspace{-0.2in}
	  \captionof{figure}{Runtime comparison of \method to \imethod w/ varying $L$ and $K$.}
	  \label{fig:scale}
	 \end{minipage}
	  \vspace{-0.1in}
\end{table}


\vspace{-0.05in}
\textbf{Q4. How sensitive hyper-ensembles are to their own HPs, which include (1) the selected HP ranges, and (2) the number of sub-models?} The two HP terms are directly related, since expanding/shrinking the HP ranges result in more/fewer sub-models constituting the ensemble.  We first answer how HP value ranges affect the accuracy of hyper-ensembles.  With MNIST-4, we expand, shrink or shift the HP ranges for the \vae , which are consitituents of \imethod.  The experimental details are provided in Appx. \ref{sssec:HP-ranges}.  As shown in Table \ref{tab:irobod-overall-result}, \imethod achieves more stable results across all settings. Moreover, \imethod produces lower variance with respect to its own HPs than that of the individual model results. 

\begin{table}[!h]
	\setlength{\tabcolsep}{5pt}
	\caption{Mean and stdev of AUROCs for \imethod vs. \vae.  \imethod's performance is evaluated with different HP ranges in Table \ref{tab:HPrange-i}.  Mean and stdev for \vae are calculated over all constituent sub-models from the same table. The hyper-ensemble \imethod has considerably lower performance variation w.r.t. its HPs, than individually traind \vae.
	}
	\vspace{-0.085in}
	\hspace{-0.0in}
	\centering
	{{
			\begin{tabular}{r@{$\pm$}l | r@{$\pm$}l}
				\toprule
				\multicolumn{2}{c}{\textbf{Mean\&Std.} (\imethod) }  & \multicolumn{2}{c}{ \textbf{Mean\&Std.} (\vae)}\\
				\midrule
				83.0&\textbf{2.8} & 78.9&9.8 \\
				\bottomrule                                                
			\end{tabular}
	}}
	\vspace{-0.07in}
	\label{tab:irobod-overall-result}
\end{table}

We also study how number of sub-models impact our model. For the same MNIST-4,  \svdd and \vae are selected as the base model for hyper-ensembles, with details in Appx. \ref{sssec:HP-numbers}. Fig. \ref{fig:submodelvsauc} shows the AUROC corresponding to different number of sub-models (for \svdd we have both Polluted (left) and Clean (middle) settings, and for \vae we have Polluted (right) setting only).  Observe that AUROC quickly stabilizes and the variance shrinks, as the number of sub-models grows beyond a certain size. The larger number of sub-models is, the more stable is the performance.  

Finally, we conduct the sensitivity analysis to both HP value ranges and the number of sub-models (since they are correlated) for our proposed \method. We evaluate on the Cardio dataset and provide the details in Appx. \ref{sssec:HP-robod}. Table \ref{tab:irobod-comparison} shows that \method yields smaller variance to its own HPs and provides more competitive results than other benchmarked models. To summarize, our experiments show that larger number of sub-models and expanded HP ranges under fine grids can relax an hyper-ensemble model's dependency on its own HPs.

\begin{figure}[!h]
	\centering
	\vspace{-0.085in}
	\hspace{-0.02in}
	\begin{tabular}{ccc}
		\includegraphics[width=0.3\textwidth]{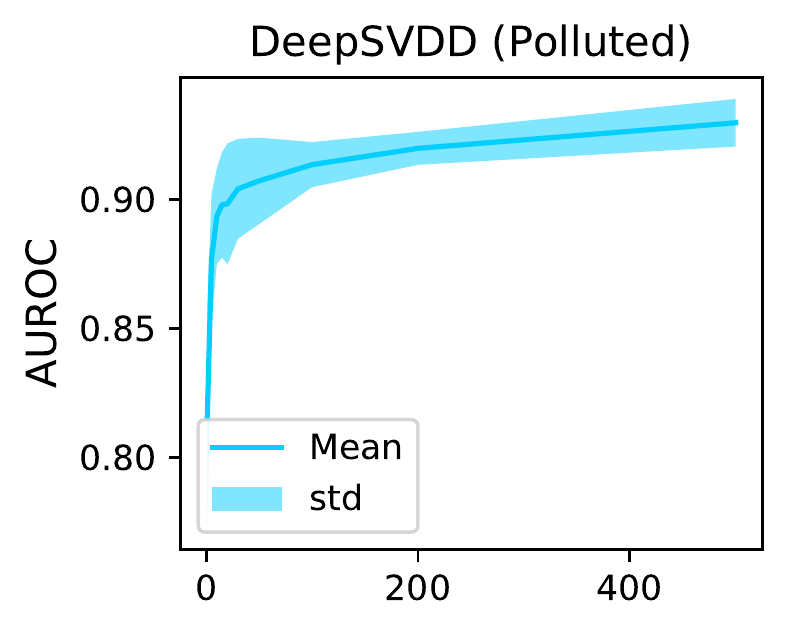} \hspace{-0.02in} \vspace{-0.035in} & 	
		\includegraphics[width=0.3\textwidth]{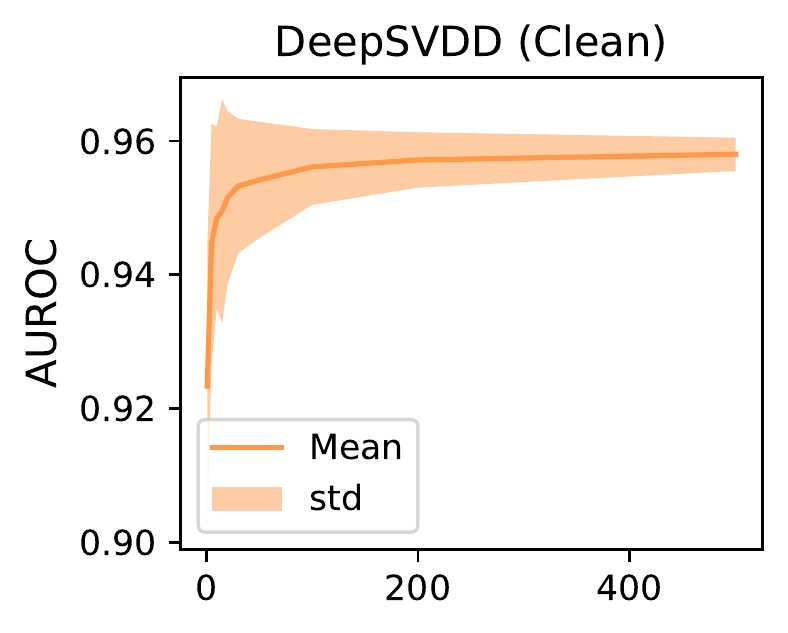} \vspace{-0.035in} \hspace{-0.035in} & 	
		\includegraphics[width=0.3\textwidth]{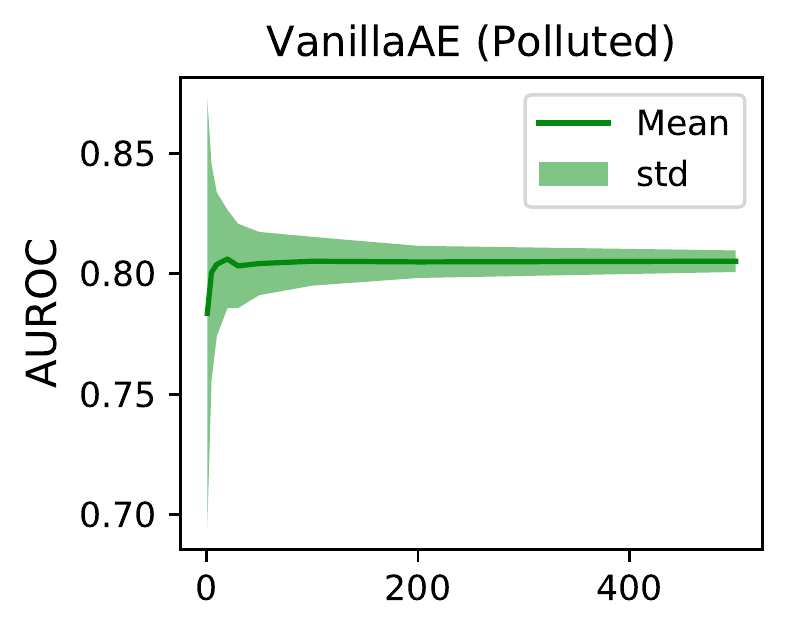} \vspace{-0.035in} \hspace{-0.035in}	\\ 
		\includegraphics[width=0.3\textwidth]{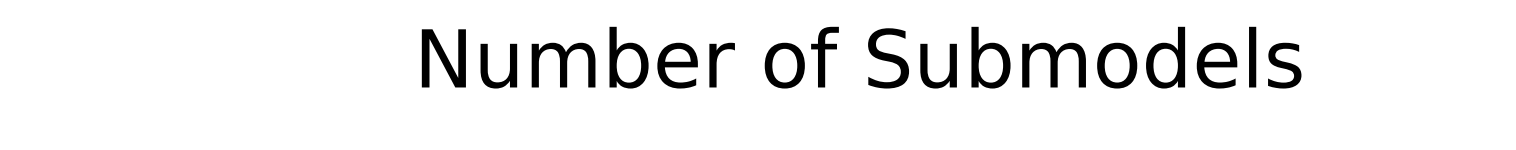} \hspace{-0.02in}&
		\includegraphics[width=0.3\textwidth]{figs/submodels.png} \hspace{-0.02in}&  
		\includegraphics[width=0.3\textwidth]{figs/submodels.png} \hspace{-0.02in}\\ 
	\end{tabular}
	\vspace{-0.25in}
	\caption{Number of sub-models vs. AUROC of ensembled-OD methods  on MNIST-4. Shaded areas depict one (1) stdev around the mean performance of the ensemble. For all cases the performance quickly improves to a stable state after $\sim$20 sub-models, with notably small standard deviation.
	}
	\label{fig:submodelvsauc}  
	\vspace{-0.06in}
\end{figure}

\begin{table}[!h]
	\setlength{\tabcolsep}{5pt}
	\caption{For Cardio dataset, we provide mean and stdev AUROCs of \method and benchmarks from HP ranges in Tables \ref{tab:ROBODrange} (\method) and \ref{tab:HP overview} (benchmarks).The hyper-ensemble \method has considerably lower performance variation w.r.t. its HPs.
	}
	\vspace{-0.07in}
	\hspace{-0.0in}
	\centering
	{{
			\begin{tabular}{r@{$\pm$}l  r@{$\pm$}l   r@{$\pm$}l  r@{$\pm$}l  r@{$\pm$}l  r@{$\pm$}l}
				\toprule
				\multicolumn{2}{c}{\method }  & \multicolumn{2}{c}{\vae} & \multicolumn{2}{c}{ \rda}& \multicolumn{2}{c}{\svdd} & \multicolumn{2}{c}{ \rand} & \multicolumn{2}{c}{ IF}\\
				\midrule
				93.6&\textbf{0.4} & 87.7&7.7 & 78.4&12.1 & 54.6&7.9 & 89.2&5.3 & 94.1&0.9  \\
				\bottomrule                                                
			\end{tabular}
	}}
	\vspace{-0.22in}
	\label{tab:irobod-comparison}
\end{table}

\section{Conclusion}
\label{sec:conclusion}

In this work, we have provided a thorough analysis on the hyperparameter (HP) sensitivitiy of several state-of-the-art deep outlier detection (OD) methods.
Our findings quantitatively confirm that model selection is vital and advocate efforts in this line of research.
To this end, we introduce \method, a scalable hyper-ensemble OD method that 
remedies the ``choice paralysis'' by assembling various autoencoder models of different HP configurations, hence obviating HP/model selection. 
We speed up ensemble training through novel strategies that simultaneously train varying depth and width models under parameter sharing.
Extensive experiments on image and point-cloud datasets show the competitiveness of \method compared to existing OD baselines, 
while providing consistent results across random initializations.
We hope that our work increases awareness to the unsupervised model selection challenge for the newly booming deep OD literature and motivates future work on hyperparameter-robust model design.

\vspace{-0.1in}


\section*{Acknowledgments}{
	This work is sponsored by NSF CAREER 1452425. We also thank PwC Risk and Regulatory Services Innovation Center at Carnegie Mellon University. Any conclusions expressed in this material are those of the author and do not necessarily reflect the views, expressed or implied, of the funding parties.
}

\clearpage
\newpage

{
\small
\bibliography{00refs}
}

\section*{Checklist}

\begin{enumerate}

\item For all authors...
\begin{enumerate}
  \item Do the main claims made in the abstract and introduction accurately reflect the paper's contributions and scope?
 \answerYes{}
  \item Did you describe the limitations of your work?
 \answerYes{} \textit{Our solution to tackle hyperparameter (HP) sensitivity for deep OD is an ensemble model. We have emphasized throughout the paper that ensemble training is costlier than training a single model. In fact, one of our key contributions is to speed up ensemble training through novel design strategies, yet, further improvements are left to future research.}
  \item Did you discuss any potential negative societal impacts of your work?
 \answerNA{}
  \item Have you read the ethics review guidelines and ensured that your paper conforms to them?
 \answerYes{}
\end{enumerate}

\item If you are including theoretical results...
\begin{enumerate}
  \item Did you state the full set of assumptions of all theoretical results?
 \answerNA{}
        \item Did you include complete proofs of all theoretical results?
 \answerNA{}
\end{enumerate}

\item If you ran experiments...
\begin{enumerate}
  \item Did you include the code, data, and instructions needed to reproduce the main experimental results (either in the supplemental material or as a URL)?
        \answerYes{} \textit{We open-source our source code (including datasets, our method as well as SOTA methods' implementations) at  \textit{{\url{https://github.com/xyvivian/ROBOD}}}, to which we point in the Introduction section. }
 \item Did you specify all the training details (e.g., data splits, hyperparameters, how they were chosen)?
        \answerYes{} \textit{Yes. We include the hyperparameters used in sensitivity analysis in   Table \ref{tab:hps} in Sec. \ref{sec:sensitive}. 
        	Additional HP information is provided in Appx. \ref{ssec:hps}.}
 \item Did you report error bars (e.g., with respect to the random seed after running experiments multiple times)?
          \answerYes{} \textit{We report empirical AUROC values across  multiple hyperparameter configurations and random initializations and report the mean as well as the standard deviation.}
 \item Did you include the total amount of compute and the type of resources used (e.g., type of GPUs, internal cluster, or cloud provider)?
         \answerYes{} \textit{We report the computational resources in Sec. \ref{sec:experiments}.}
\end{enumerate}

\item If you are using existing assets (e.g., code, data, models) or curating/releasing new assets...
\begin{enumerate}
  \item If your work uses existing assets, did you cite the creators?
   	\answerYes{} \textit{We utilize standard image datasets (MNIST and CIFAR10) along with 3 tabular dataset. We reference the public repository from which the tabular 
   	 data are collected.}
  \item Did you mention the license of the assets?
   	\answerNo{} \textit{All the datasets we experimented with are open-source with a public license.}
  \item Did you include any new assets either in the supplemental material or as a URL?
  	\answerNo{}
  \item Did you discuss whether and how consent was obtained from people whose data you're using/curating?
    \answerNA{}
  \item Did you discuss whether the data you are using/curating contains personally identifiable information or offensive content?
    \answerNA{}
\end{enumerate}

\item If you used crowdsourcing or conducted research with human subjects...
\begin{enumerate}
  \item Did you include the full text of instructions given to participants and screenshots, if applicable?
     \answerNA{}  
  \item Did you describe any potential participant risks, with links to Institutional Review Board (IRB) approvals, if applicable?
    \answerNA{}
  \item Did you include the estimated hourly wage paid to participants and the total amount spent on participant compensation?
    \answerNA{}
\end{enumerate}

\end{enumerate}


\clearpage 
\appendix

\section{Appendix}
\label{sec:appx}

\subsection{Preview of Existing Deep OD Methods}
\label{ssec:existing}

\setlength{\tabcolsep}{4pt}
\begin{table}[h]
\caption{
Representative unsupervised deep OD models from 4 different families (for a broader coverage, see surveys \cite{pang2021deep,UnifyingSurvey,chalapathy2019learning}), annotated in terms of data used for training, test, and validation/model selection (if any).
No existing work attempts (unsupervised) model selection; vast majority reports results for a \textit{fixed} (how, unclear) ``recommended''  config. or tune {\em some} but not all HPs using labeled validation and even test (!) data. 
AE: autoencoder, SSL: self-supervised learning, Clean: inlier-only.
\label{tab:unsupmethods}
}
\vspace{-0.05in}
\hspace{-0.1in}
{\scalebox{0.785}{
\begin{tabular}{p{1.2in}cl|lll}
\toprule
\textbf{Method} & {\bf Year} 
& {\bf Family}
& \textbf{Train}
&  \textbf{Test}
& \textbf{Validation (HP/Model Selection)}    \\
\midrule
RandNet \cite{conf/sdm/ChenSAT17} & 2017 & AE ensemble & Polluted  & =Train & None, fixed -- sensitivity analysis on some HPs \\
RDA \cite{conf/kdd/ZhouP17} & 2017 & AE & Polluted  &  =Train  & Best $\lambda$ on Test, other HPs fixed \\
DAGMM \cite{conf/iclr/ZongSMCLCC18} & 2018 & AE \& density & Clean \& Pol.d & Disjoint  & None, fixed -- sensitivity on reg. param.s $\{\lambda_1, \lambda_2\}$ \\ 
\hline 
DeepSVDD \cite{conf/icml/RuffGDSVBMK18} & 2018 & One-Class & Clean & Disjoint   & Best $\nu$ on Test, other HPs fixed \\
DROCC \cite{goyal2020drocc} & 2020 & One-Class &   Clean & Disjoint & Validation data to tune some (not all) HPs \\
HRN \cite{hu2020hrn} & 2020 & One-Class &   Clean & Disjoint & 10\% of Test for tuning $\{\lambda, n\}$, other HPs fixed \\
\hline 
(f-)AnoGAN \cite{conf/ipmi/SchleglSWSL17,journals/mia/SchleglSWLS19} & 2017 & GAN & Clean \& Pol.d & Disjoint  & None, fixed \\ 
EGBAD \cite{zenati2018efficient} & 2018 & (Bi)GAN &  Clean & Disjoint  &  None, fixed \\ 
GANomaly \cite{akcay2018ganomaly} & 2018 & GAN & Clean & Disjoint  & Best reg. weights $\{w_{adv},w_{con},w_{enc}\}$ on Test, others fixed \\ 

\hline 
GOAD \cite{bergman2020classificationbased} & 2020 & SSL &  Clean & Disjoint  & None, fixed \\
NeuTraL \cite{boyd2020neural} & 2020 & SSL & Clean & Disjoint  & 10\% of Test for tuning transformation HPs, others fixed  \\
 \bottomrule
\end{tabular}
}}
\vspace{-0.15in}
\end{table}

\subsection{Details on Hyperparameter-Sensitivity Analysis}
\label{ssec:hps}

{\bf Clean versus Polluted Testbed Setup.~}
For sensitivity analysis, we construct our testbed on several datasets, including MNIST, CIFAR10, Thyroid and Cardio. For MNIST, we choose Digit `4' and `5' as the inlier-class, individually. For CIFAR10, we choose class `automobile' as the inlier-class. The inlier-class is assigned the label 0, while we regard all classes other than the inlier-class as outliers and mark them with label 1 instead. Since MNIST and CIFAR10 are image data, we first apply the global contrast normalization to each individual image. We utilize the default train/test data-split (supported with Pytorch vision package). In the Clean setting, we select only the inlier-class data from train data-split as the training data. We measure and report the AUROC of the compared methods on the test data-split, with label 0 being the inlier-class and 1 being rest of the classes. In the Polluted setting, we utilize all the inlier-class data from the train data-split as the label 0, and we mix the data with $10\%$ outlier-classes' data within the train data-split. 

The tabular data, Thyroid and Cardio, are downloaded from the ODDS repository (available at \url{http://odds.cs.stonybrook.edu/}. The data come in Polluted setting, with $2.5\%$ and $9.6\%$ outliers, respectively. The inliers are labeled 0 and outliers are denoted as label 1. For tabular data, we transform and scale each feature between zero and one.

We also conduct experiment using \gan \cite{akcay2018ganomaly}'s data (MNIST digit `4' as the outlier class, rest as inliers). The data split and configuration are the same as described in the authors' provided code.

{\bf Model HP Descriptions and Grid of Values.~}

\cbit
\item \vae: 
\cben
\item {\tt n\_layers}: number of encoder layers (i.e. depth)
\item {\tt layer\_decay}: the rate of NN width's shrinkage between current and next encoder layers, the decoder layers are expanded at the same rate.
\item {\tt LR}: learning rate
\item {\tt iter}: number of epochs/iterations
\ceen

\item \rda: 
\cben
\item $\lambda$ (model-specific reg.): a penalty term that tunes the level of sparsity in the outlier matrix $S$ (refer to Section 3.1 in \cite{conf/kdd/ZhouP17}).
\item {\tt n\_layers}: number of encoder layers (i.e. depth)
\item {\tt layer\_decay}: the rate of NN width's shrinkage between current and next encoder layers, the decoder layers are expanded at the same rate.
\item {\tt LR}: learning rate
\item {\tt inner\_iter}: number of epochs/iterations to train the underlying autoencoder (AE), before updating the outlier matrix $S$ and inlier matrix $L$ (refer to Section 4.1 in \cite{conf/kdd/ZhouP17}).
\item {\tt iter}: number of epochs/iterations in the algorithm, which first separates the training data into outlier matrix $S$ and inlier matrix $L$, then trains AE on the inlier matrix. 
\ceen

\item \svdd:
\cben
\item {\tt conv\_dim}: the output number of channels, after the first- convolutional encoder layer. After the first-layer, the number of channels expand at rate of $2$.
\item {\tt fc\_dim}: the output dimension of the fully connected layer between convolutional encoder layers and decoder layers, in the LeNet structure \cite{lecun1998gradient}.
\item {\tt Relu\_slope}: \svdd utilizes leaky-relu activation to avoid the trivial, uninformative solutions \cite{conf/icml/RuffGDSVBMK18}. Here we alter the leakiness of the relu sloping.
\item {\tt pretr\_iter}: In \cite{conf/icml/RuffGDSVBMK18}, an AE is pre-trained, to set the hypersphere center $c$ to the mean of the
mapped data. pretr\_iter determines the number of epochs/iterations to train AE.
\item {\tt pretr\_LR}: the learning rate to pretrain the AE.
\item {\tt iter}: the number of epochs/iterations in training the \svdd
\item {\tt LR}: the learning rate during training
\item {\tt wght\_dc}: weight decay rate
\ceen

\item \gan
\cben
\item $w_{adv}$: weight parameter that adjusts the adversarial loss function (See Section 3.2 in \cite{akcay2018ganomaly}.)
\item $w_{con}$: weight parameter that adjusts the contextual loss, for learning the contextual information about the
input data (See Section 3.2 in \cite{akcay2018ganomaly}.)
\item $w_{enc}$: weight parameter that adjusts the encoder loss and minimizes the distance between the
bottleneck features and the encoder of the generated features (See Section 3.2 in \cite{akcay2018ganomaly}.) 
\item {\tt z\_dim}: the dimension of the reduced embedded space after the input is passed through the encoders 
\item {\tt LR}: learning rate
\item {\tt iter}: number of epochs/iterations
\ceen

\item \rand 
\cben
\item {\tt n\_layers}: number of encoder layers (i.e. depth) in BAE
\item {\tt layer\_decay}: the rate of NN width's shrinkage between current and next encoder layers, the decoder layers are expanded at the same rate.
\item {\tt sample\_r}: sample size selection for adaptive sampling (See Section 3.3 in \cite{aggarwal2017outlier}.)
\item {\tt ens\_size}: number of ensemble members/models 
\item {\tt pretr\_iter}: number of epochs/iterations to pretrain the BAE
\item {\tt iter}: number of epochs/iterations
\item {\tt LR}: learning rate
\item {\tt wght\_dc}: weight decay rate
\ceen
\ceit

\begin{table}[]
		\caption{
		We define a grid of 1-3 unique values for each hyperparameter (HP) of each deep OD method studied.
		With 4-to-8 different HPs each, the total number of configurations, and i.e. models trained, quickly grows to several hundreds.
		When applicable/available, we include the \textbf{\underline{author-recommended}} value (marked in bold and underlined) in the respective grid.
			\label{tab:grids}
		}
		\vspace{-0.15in}
		\hspace{-0.1in}
		\centering
		{\scalebox{0.8}{
	\begin{tabular}{l|lllllll}
		\hline
		\multicolumn{1}{|l|}{\textbf{Method}}             & \multicolumn{1}{l|}{\textbf{Hyperparameter}} & \multicolumn{1}{l|}{\textbf{Grid}}    & \multicolumn{1}{l|}{\textbf{\#values}} & \multicolumn{1}{l|}{\textbf{Method}}           & \multicolumn{1}{l|}{\textbf{Hyperparameter}} & \multicolumn{1}{l|}{\textbf{Grid}}    & \multicolumn{1}{l|}{\textbf{\#values}} \\ \hline
		\multicolumn{1}{|l|}{\multirow{4}{*}{\vae}} & n\_layers                                    & {[}2, 3, 4{]}                           & \multicolumn{1}{l|}{3}                 & \multicolumn{1}{l|}{\multirow{8}{*}{\svdd}} & conv\_dim                                    & {[}\textbf{\underline{8}}, 16, 32{]}                         & 3                                      \\
		\multicolumn{1}{|l|}{}                            & layer\_decay                                 & {[}1, 2, 4{]}                           & \multicolumn{1}{l|}{3}                 & \multicolumn{1}{l|}{}                          & fc\_dim                                      & {[}\textbf{\underline{16}}, 32{]}                           & 2                                      \\
		\multicolumn{1}{|l|}{}                            & LR                                           & {[}1e-3, 1e-4, 1e-5{]}                 & \multicolumn{1}{l|}{3}                 & \multicolumn{1}{l|}{}                          & Relu\_slope                                  & {[}\textbf{\underline{1e-1}}, 1e-3{]}                       & 2                                      \\
		\multicolumn{1}{|l|}{}                            & iter                                         & {[}200, 500, 1000{]}                    & \multicolumn{1}{l|}{3}                 & \multicolumn{1}{l|}{}                          & pretr\_iter                                  & {[}200, \textbf{\underline{350}}, 400{]}                     & 3                                      \\ \cline{1-4}
		& \multicolumn{1}{l|}{}                        & \multicolumn{1}{l|}{Total =}          & \multicolumn{1}{l|}{81}                & \multicolumn{1}{l|}{}                          & pretr\_LR                                    & {[}1e-4, 1e-5{]}                      & 2                                      \\ \cline{1-4}
		\multicolumn{1}{|l|}{\multirow{6}{*}{\rda}}        & $\lambda$                                    & {[}5e-1, 5e-3, 5e-5{]}                  & \multicolumn{1}{l|}{3}                 & \multicolumn{1}{l|}{}                          & iter                                         & {[}100, 200, \textbf{\underline{250}}{]}                     & 3                                      \\
		\multicolumn{1}{|l|}{}                            & n\_layers                                    & {[}2, 3, 4{]}                           & \multicolumn{1}{l|}{3}                 & \multicolumn{1}{l|}{}                          & LR                                           & {[}1e-4, 1e-5{]}                      & 2                                      \\
		\multicolumn{1}{|l|}{}                            & layer\_decay                                 & {[}1, 2, 4{]}                           & \multicolumn{1}{l|}{3}                 & \multicolumn{1}{l|}{}                          & wght\_dc                                     & {[}1e-5, \textbf{\underline{1e-6}}{]}                      & 2                                      \\ \cline{5-8} 
		\multicolumn{1}{|l|}{}                            & LR                                           & {[}1e-3, 1e-4{]}                      & 2                                      & \multicolumn{1}{l|}{}                          & \multicolumn{1}{l|}{}                        & \multicolumn{1}{l|}{Total \#models =} & \multicolumn{1}{l|}{864}               \\ \cline{5-8} 
		\multicolumn{1}{|l|}{}                            & inner\_iter                                  & {[}20, 50{]}                           & \multicolumn{1}{l|}{2}                 & \multicolumn{1}{l|}{\multirow{8}{*}{\rand}}  & n\_layers                                    & {[}3, 5, \textbf{\underline{7}}, 9{]}                         & 4                                      \\
		\multicolumn{1}{|l|}{}                            & iter                                         & {[}5, 20, 50{]}                         & \multicolumn{1}{l|}{3}                 & \multicolumn{1}{l|}{}                          & layer\_decay                                 & {[}0.3, \textbf{\underline{0.6}}{]}                        & 2                                      \\ \cline{1-4}
		& \multicolumn{1}{l|}{}                        & \multicolumn{1}{l|}{Total \#models =} & \multicolumn{1}{l|}{324}               & \multicolumn{1}{l|}{}                          & sample\_r                                    & {[}1.00, \textbf{\underline{1.01}}{]}                      & 2                                      \\ \cline{1-4}
		\multicolumn{1}{|l|}{\multirow{6}{*}{\gan}}   & $w_{adv}$                                    & \textbf{\underline{1}}                                     & \multicolumn{1}{l|}{1}                 & \multicolumn{1}{l|}{}                          & ens\_size                                    & {[}\textbf{\underline{50}}, 200{]}                          & 2                                      \\
		\multicolumn{1}{|l|}{}                            & $w_{con}$                                    & {[}25, \textbf{\underline{50}}, 100{]}                       & \multicolumn{1}{l|}{3}                 & \multicolumn{1}{l|}{}                          & pretr\_iter                                  & 100                                   & 1                                      \\
		\multicolumn{1}{|l|}{}                            & $w_{enc}$                                    & {[}0.1, \textbf{\underline{1}}{]}                          & \multicolumn{1}{l|}{2}                 & \multicolumn{1}{l|}{}                          & iter                                         & {[}\textbf{\underline{300}}, 1000{]}                       & 2                                      \\
		\multicolumn{1}{|l|}{}                            & z\_dim                                       & {[}50, \textbf{\underline{100}}, 200{]}                      & \multicolumn{1}{l|}{3}                 & \multicolumn{1}{l|}{}                          & LR                                           & {[}1e-2, 1e-3, 1e-4{]}                & 3                                      \\
		\multicolumn{1}{|l|}{}                            & LR                                           & {[}5e-3, \textbf{\underline{2e-3}}, 5e-4{]}                 & \multicolumn{1}{l|}{3}                 & \multicolumn{1}{l|}{}                          & wght\_dc                                     & 0                                     & 1                                      \\ \cline{5-8} 
		\multicolumn{1}{|l|}{}                            & iter                                         & {[}10, \textbf{\underline{15}}, 25{]}                        & 3                                      & \multicolumn{1}{l|}{}                          & \multicolumn{1}{l|}{}                        & \multicolumn{1}{l|}{Total \#models =} & \multicolumn{1}{l|}{192}               \\ \cline{1-4} \cline{6-8} 
		& \multicolumn{1}{l|}{}                        & \multicolumn{1}{l|}{Total \#models =} & \multicolumn{1}{l|}{162}               &                                                &                                              &                                       &                                        \\ \cline{2-4}
	\end{tabular}
}}
\end{table}

\subsection{Additional Results: Hyperparameter-Sensitivity Analysis}
\label{ssec:hpa}

In Fig. \ref{fig:mnist4Clean}, we show the AUROC performance of deep OD methods in Clean setting.  In  Fig. \ref{fig:mnist4in}, Fig. \ref{fig:mnist5inCIFARauto}, Fig. \ref{fig:ganomaly} and Fig. \ref{fig:randnet} we show additional experiment results and AUROC performances over $3$ runs with different random initializations. 

\begin{figure}[!h]
	\vspace{-0.1in}
	\centering
	\begin{tabular}{c}
		\includegraphics[width=0.7\textwidth, height=1.75 in]{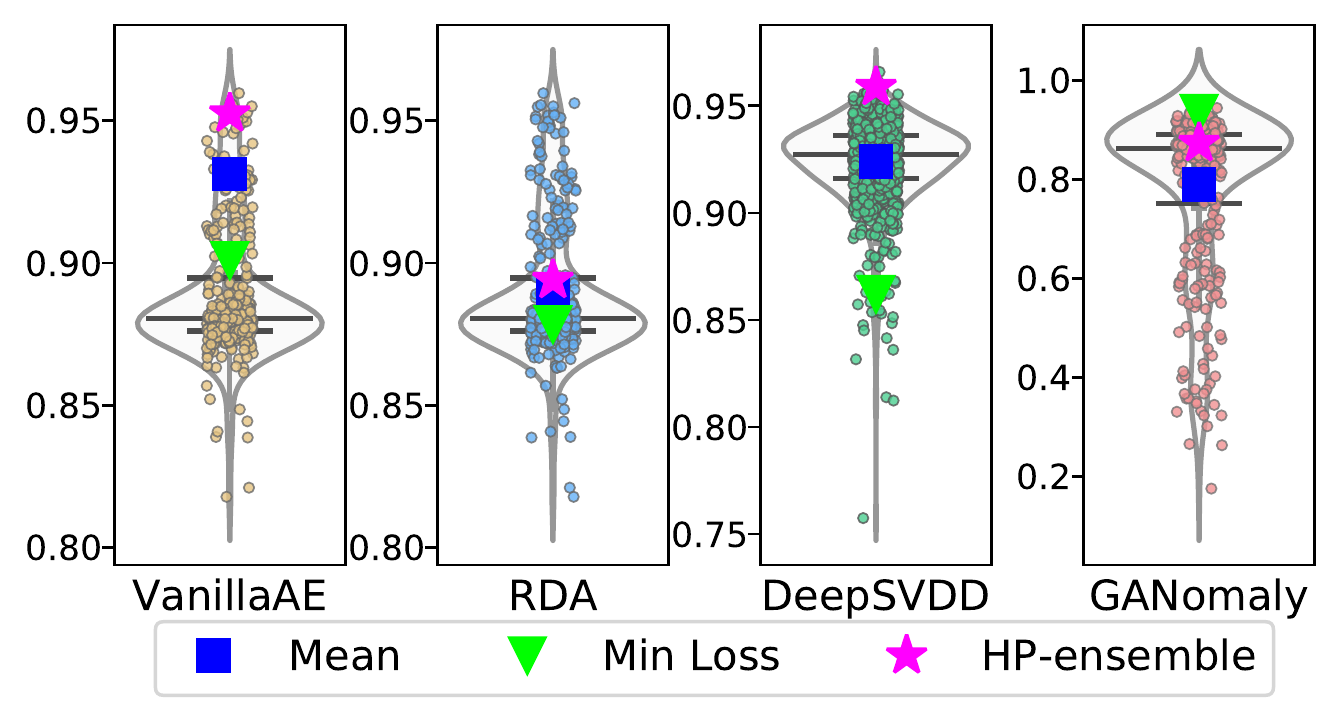} 
	\end{tabular}
	 \vspace{-0.05in}
	\caption{\label{fig:mnist4Clean} AUROC performance of deep OD methods with different HP configurations (circles) on MNIST-4 with Clean training data showcase notable variation (i.e., sensitivity). Hyper-ensemble (\FiveStarOpen) improves notably over Mean ($\square$). }
	\vspace{-0.2in}
\end{figure}

\begin{figure}[!ht]
\centering
\begin{tabular}{c}
	\includegraphics[width=\textwidth]{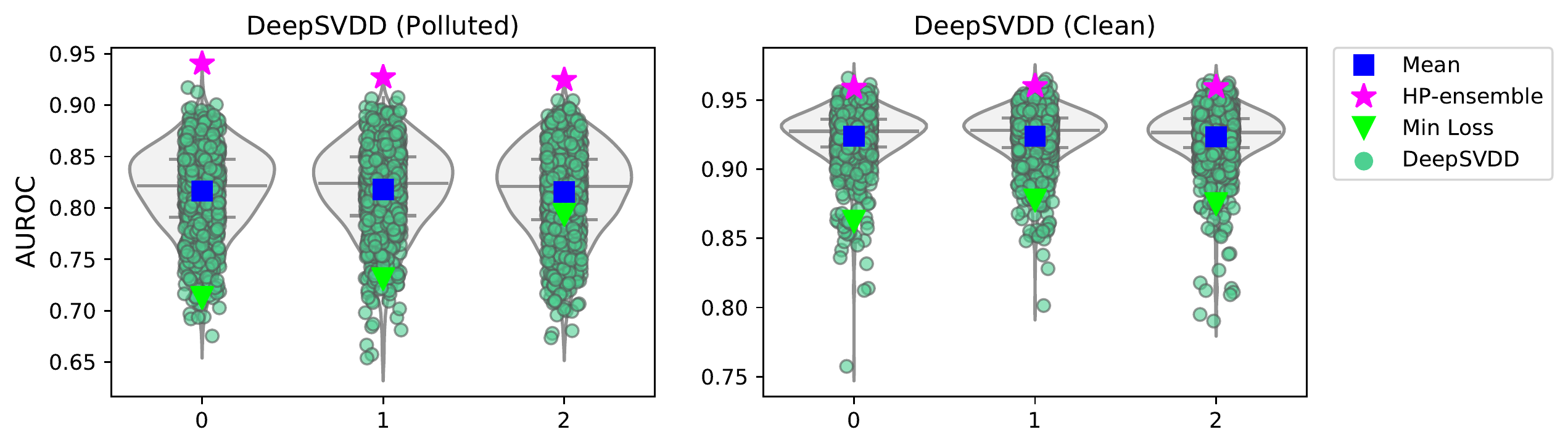}  \\
	\includegraphics[width=\textwidth]{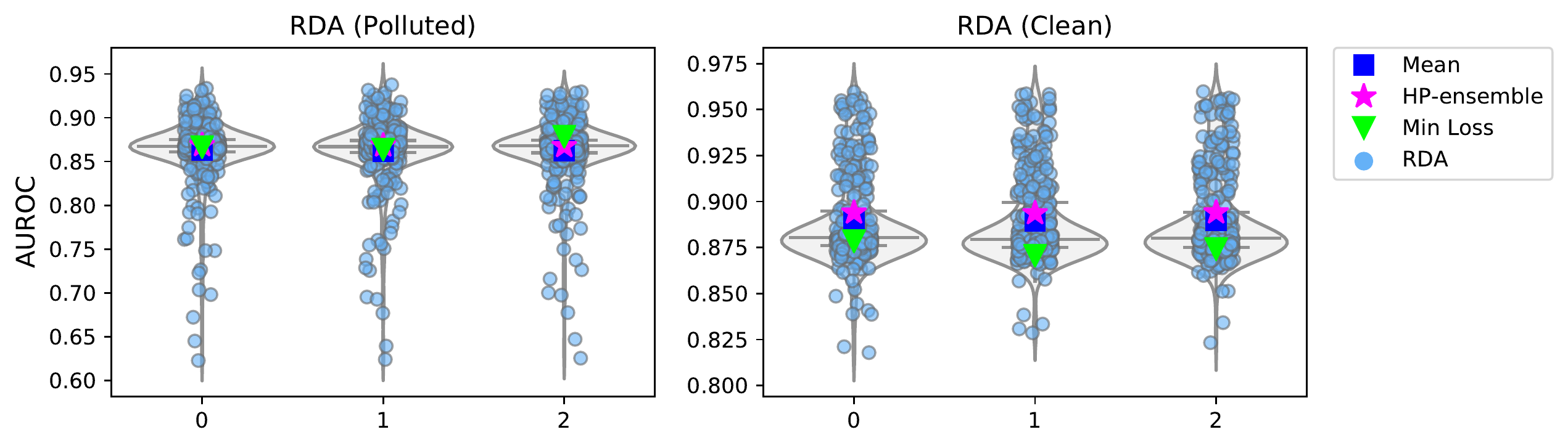} \\
	\includegraphics[width=\textwidth]{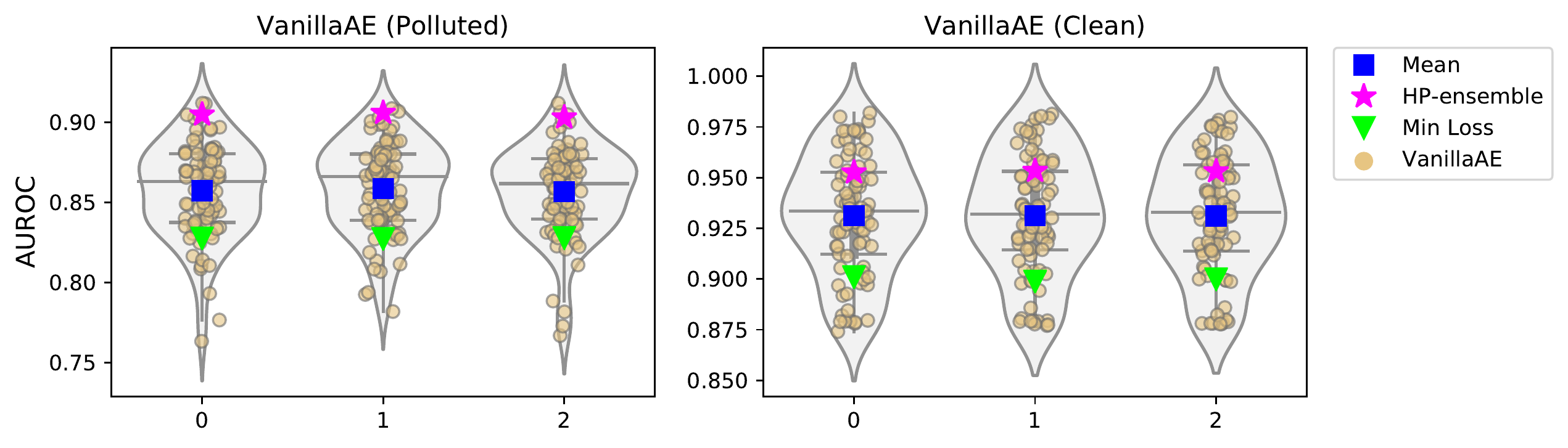} \\	
   	\includegraphics[width=\textwidth]{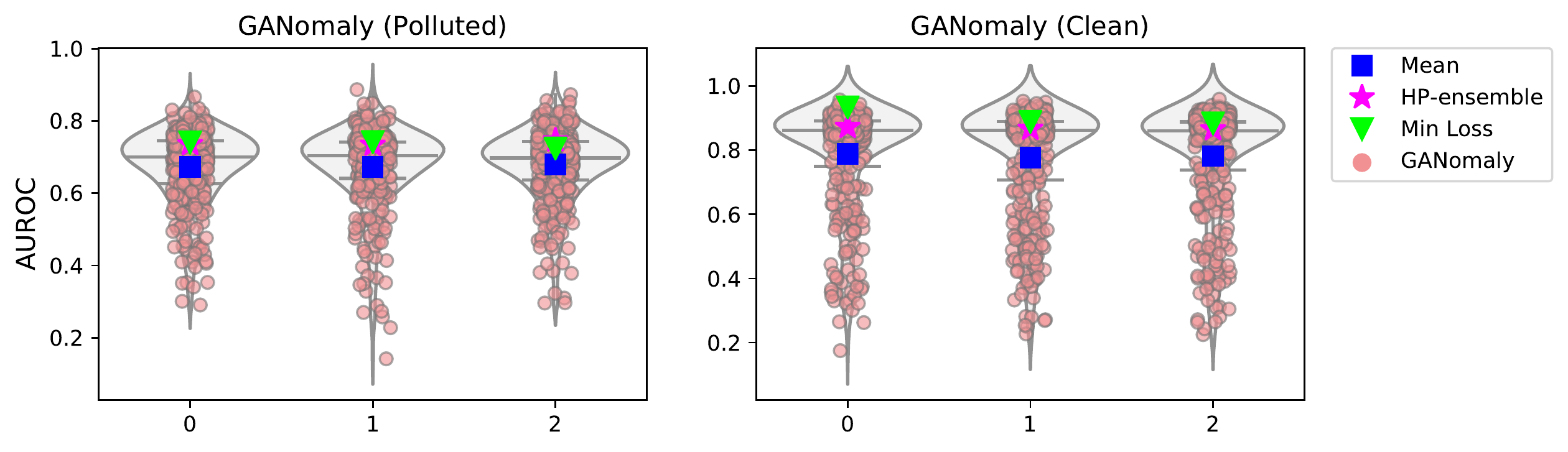}\\
\end{tabular}
 \vspace{-0.2in}
	\caption{\label{fig:mnist4in} AUROC results under varying HP-configurations on MNIST-4 dataset under the Clean (only train on digit `4' images) and Polluted (digit `4' as inliers,  the rest nine classes down-sampled at 10\% as outliers) settings. We conduct each experiment 3 times with different random initialization, where each plot's x-axis corresponds to the experiment index. Note: y-axes are not directly comparable -- we use different y-axis to better reflect the spread for each experiment.}
\end{figure}

\begin{figure}[!ht]
	\centering
	\begin{tabular}{c}
		\includegraphics[width=\textwidth]{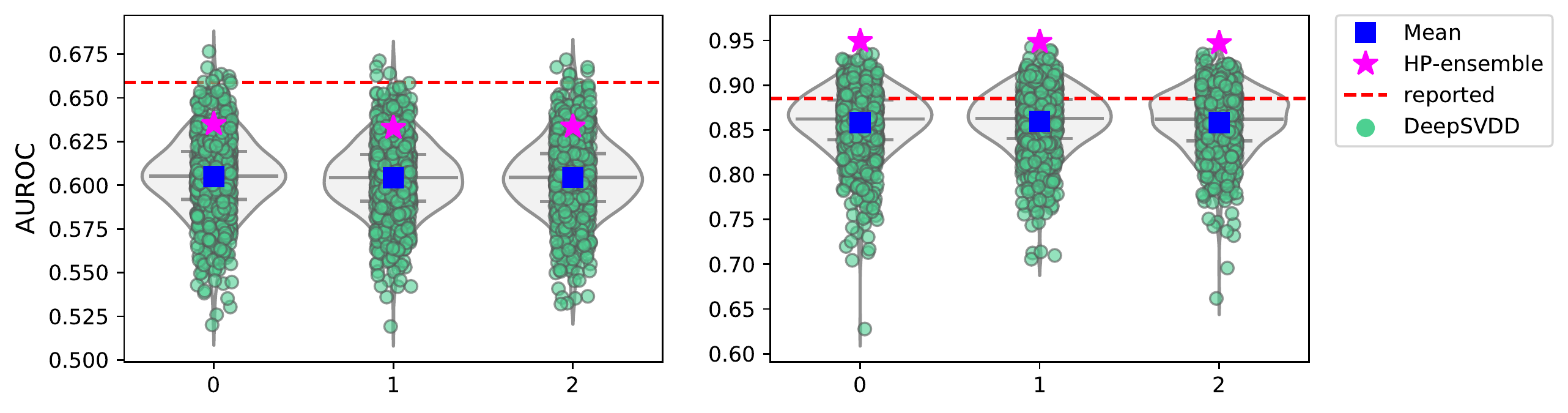} \\ 
\end{tabular}
\vspace{-0.2in}
	\caption{\label{fig:mnist5inCIFARauto} For \svdd algorithm, we show various HP-configurations' AUROC results under the Clean setting. Left: MNIST digit `5' as the inlier data. Right: CIFAR10 `automobile' as the inlier data. We conduct each experiment 3 times with different random initialization, where each plot's x-axis corresponds to the experiment index. Performance reported in the original paper \cite{conf/icml/RuffGDSVBMK18} appears above what we have obtained on average for both datasets.}
	\vspace{-0.1in}
\end{figure}

\begin{figure}[!ht]
	\centering
		\includegraphics[width=0.62\textwidth]{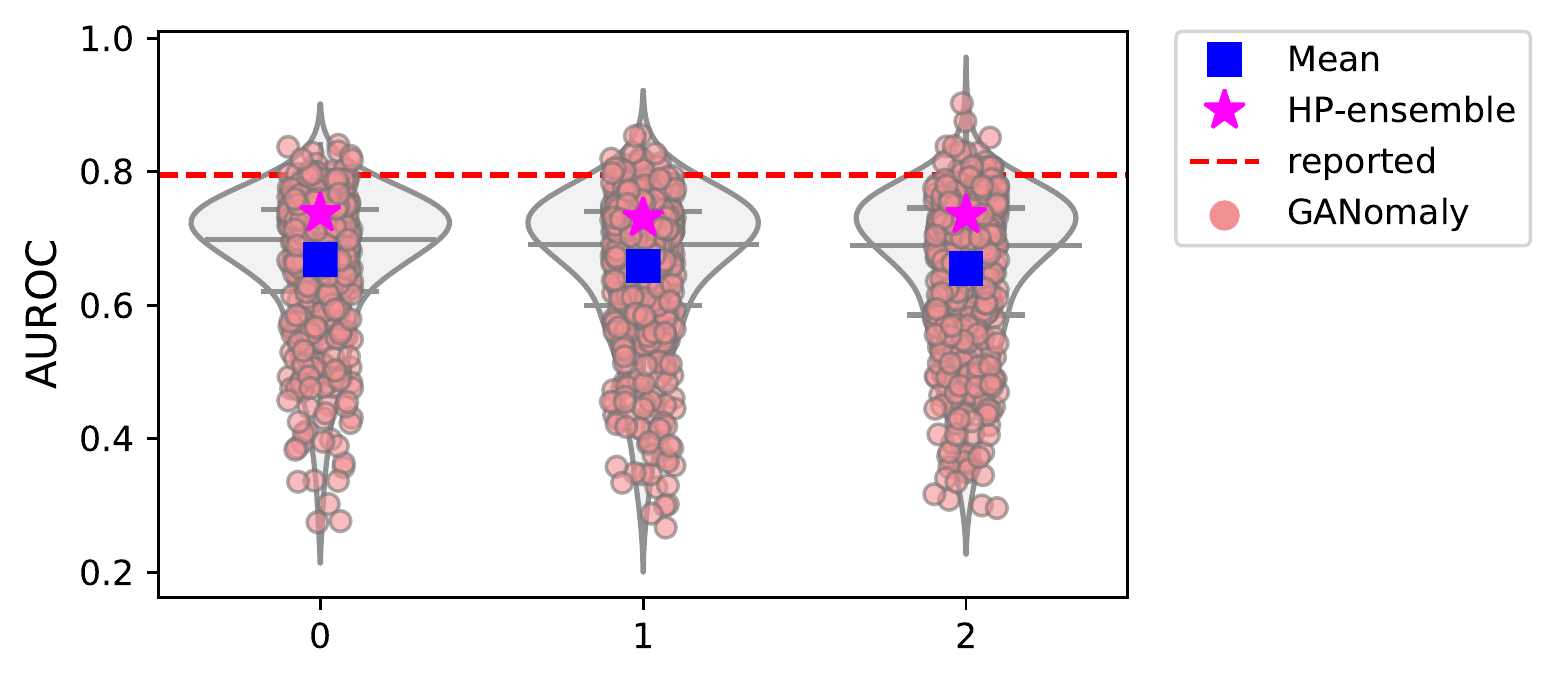} 
	\vspace{-0.2in}
	\caption{\label{fig:ganomaly} AUROC results with varying HP configurations for the \gan algorithm. We utilize the same experimental setting as in \cite{akcay2018ganomaly}, with MNIST digit `4' as the outlier class, and rest of the digit classes as inliers (also in Clean setting). The x-axis corresponds to the experiment index among 3 independent runs each with a different random initialization. Performance reported in the original paper \cite{akcay2018ganomaly} appears above what we have obtained on average.}
	\vspace{-0.1in}
\end{figure}

\begin{figure}[!ht]
	\centering
	\begin{tabular}{c}
		\includegraphics[width=\textwidth]{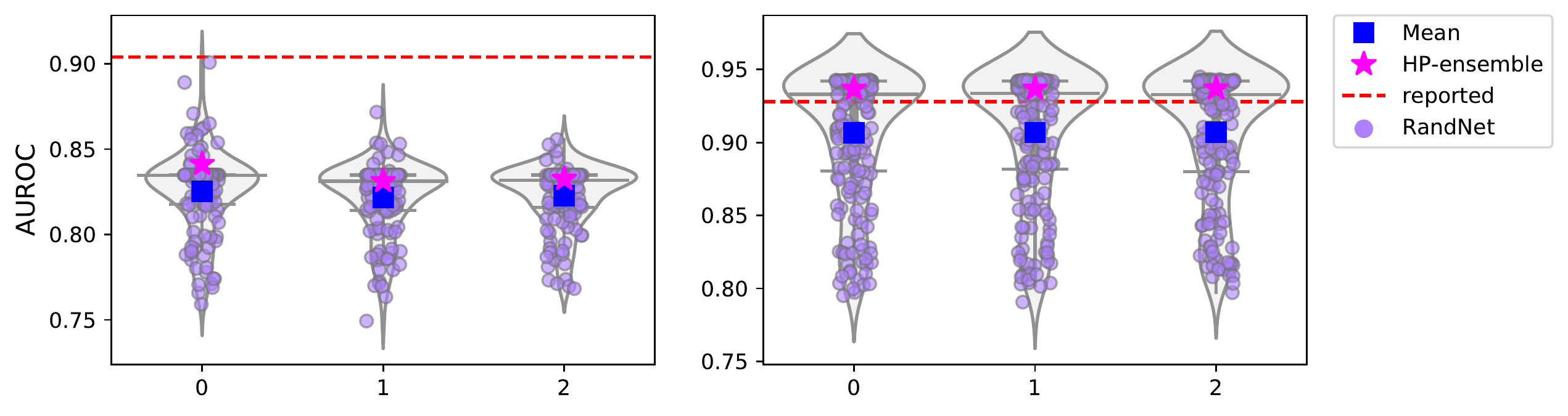} \\ 
\end{tabular}
 \vspace{-0.2in}
	\caption{\label{fig:randnet} AUROC results with varying HP configurations for the \rand algorithm. As RandNet implementation contains fully connected layers, we evaluate it only on tabular data under Polluted setting, similar to the experiments in \cite{aggarwal2017outlier}. Left: Thyorid dataset, Right: Cardio dataset. Each plot's x-axis corresponds to the experiment index for 3 different runs with random initialization.}
	\vspace{-0.1in}
\end{figure}

\subsection{Regularization Effect of Weight Sharing}
\label{ssec:reg}
One noticeable problem with applying autoencoders (AE) and reconstruction loss for outlier detection is that for AE, the decision boundary is hard to draw due to noise and outliers that can impact the quality of the reconstruction.
 In some cases, the AE
 can overfit to all the data points including the outliers, causing large false negative rate. In other cases, it may underfit to data and fail to reconstruct the input satisfactorily. While the ``denoising AEs'' \cite{xie2012image} or ``correntropy AEs'' \cite{wang2014robust} may help to alleviate such problems, they both rely on \textit{clean} inlier-only data, which is typically not available in  real-world scenarios. On the other hand, \method with \aes allows an implicit ensemble of various NN depths and widths and plays a regularization effect on the outliers scores, in effect helping prevent AE from potential failures due to underfitting or overfitting.

To demonstrate the regularization effect with \aes, we compare the reconstructed images of AE versus \aes, with number of layers $\{2,4,6,8,10,12,14\}$, respectively (hidden dimension decays at a constant rate of $2$). We compare the individual AE and \aes, keeping the other HP-configurations same, and training under the Polluted setting with MNIST digit `5' as the inlier class. Fig. \ref{fig:mnist5examples} shows the reconstructed images with these $7$ individual AEs and a single \aes ensemble, with implicit \aei ($i \in \{2\ldots 14\}$). 
We see that individual AEs are overfitting to the outlier classes (digit `3' and `7') providing  good reconstructions when layers $L$ equal to $2$ and $4$. In contrast, they underfit and fail to reconstruct any inliers or outliers when layers $L$ equals to $6$, $8$, and $10$. When $L$ increases to $12$ and $14$ layers, individual AEs provide low-quality reconstructions to both inliers and outliers, distorting all outliers to inlier class (digit `5'). In contrast, \aes provides lower-resolution reconstruction for outlier classes during \textit{AE-2} and \textit{AE-4}, that is overfitting showcases at a lower degree. \aes can still provide signal to distinguish the outliers from the inliers from \textit{AE-6} through \textit{AE-12} where the outliers gradually become more blurred and start to deform into inlier's shape. Only at very large depth at \textit{AE-14},  \aes cannot distinguish between outlier instances from inlier instances, providing low quality predictions to both classes. 

Intuitively, this regularization effect is due to the weight sharing between \aei's. Since the next \aei utilizes the weights optimized by the previous \aei, the training phase becomes easier and underfitting is less likely to occur. Moreover, \aes has a similar structure as U-Net \cite{unet}, which is known to reduce the overfitting in  medical image segmentation tasks. 

While both \imethod and \method average the reconstruction loss from ensemble members, \method with \aes structure (and hence parameter sharing) is able to better capture the outlier information thanks to this regularization effect. This phenomenon also explains why \method's performance is better than that of \imethod, e.g. on MNIST datasets in Table \ref{tab:AUC result}.

\subsection{Details on Experiment Setup}
\label{ssec:setdetails}

\subsubsection{Hyperparameter Configurations: Details}
\label{sssec:HPdetails}
In experiments, we compare to \vae, \rda \cite{conf/kdd/ZhouP17}, \svdd \cite{conf/icml/RuffGDSVBMK18} and \rand \cite{conf/sdm/ChenSAT17}. We have not compared to \gan due to the higher variance of performances we observed during sensitivity analysis. We define a small grid of values for the HPs of each of these methods. 

Because \svdd is originally trained with LeNet \cite{lecun1998gradient} (Convolutional AE) structure, we also implement Convolutional AEs for algorithms that are either pretrained with AEs, or utilize AEs as the backbone algorithm. The detailed HP configurations are shown in Table \ref{tab:HP overview}. 
The \vae and \rand are trained with AE for all three kinds of datasets; \svdd is trained with Convolutional AE (LeNet) on image data and AE on tabular data; \rda is trained on AE for MNIST and tabular data, it utilizes Convolutional AE (LeNet) on CIFAR10. If an algorithm is trained with AE as the underlying structure, we define a shared grid of HPs:  number of encoder layers, decay rate (the rate of NN width's shrinkage between current and next encoder layers), dropout rate, train learning rate, etc. Similarly for Convolutional AE, the algorithms apply the same grid of HPs: convolution channels, fully-connected layer dimensions, weight decay, learning rate, etc.

With respect to model-specific HPs, \rda uses $\lambda$ as a penalty constant for sparsity of the outlier matrix. It also uses \texttt{inner\_iters} and \texttt{iter}, to specify the number of epochs respectively  for training an AE and for separating the data into outlier and inlier matrices. For \rand, we fix the number of ensemble members to $5$ due to computational overhead in training high-dimensional image datasets, while \cite{aggarwal2017outlier} uses $50$  ensemble members  as \rand is trained on tabular data only. Other model-specific HP descriptions can be found in Sec. \ref{ssec:hps} Model HP Descriptions and Grid of Values.

For each data point, i-\method provides a score by averaging each individual \vae's reconstruction error for a data point, thus the training time required for i-\method sums up each \vae's time. \method speeds up i-\method with fast ensembles across varying NN depths and widths. Table \ref{tab:HPROBODoverview} shows the architecture overview for \method. Specifically, the NN depths and widths are set to $8$ and $6$, representing the implicit ensembles. \method explicitly ensembles over various train iterations, learning rates, dropout rate, weight decay, providing an averaged reconstruction error score for each data point. We also experiment with two subsampling based versions, denoted \method-$\delta$, where $\delta = 0.1$ and $0.5$, respectively. We let \method-$\delta$ to train each ensemble member only on $10\%$ or $50\%$ of the training data, and score ``out-of-sample'' points, i.e. the rest of the (unseen) data points.

\begin{table}[!t]
    \small
    \centering
    \caption{Grid of values for the HPs and neural architectures used in experiments.}
    \begin{tabular}{lrrrrr}
        \toprule
         HPs & AE      & LeNet (MNIST) & LeNet (CIFAR10)   \\
        \midrule
        Number of encoder layers & [2,3,4,5,6] & [2] & [3]  \\
        Decay rate   & [1.5,1.75,2,2.25,2.5,2.75,3,3.25]       & - &   -   \\
        Convolution channels & -  & [8] & [16] \\
        FC layer dimensions  & - & [16,32,64] & [32,64,128]\\ 
        Dropout rate & [0.0,0.2]  & - & - \\
        Weight decay & [0,1e-5]  & [0,1e-5,1e-6] & [0,1e-5,1e-6] \\
        Train Learning Rate & [1e-3,1e-4] & [1e-4,1e-5] & [1e-4,1e-5] \\
        \bottomrule
    \end{tabular}
    \begin{tabular}{lrrr}
    \toprule
    NN settings for dataset: & MNIST & CIFAR10 & Tabular Data \\
    \midrule
    \vae & AE & AE & AE \\
    \svdd  & LeNet (MNIST) & LeNet (CIFAR10) & AE\\
    \rda  & AE & LeNet (CIFAR10) & AE\\
    \rand & AE & AE & AE \\
    \bottomrule
    \end{tabular}
    
    \begin{tabular}{lr}
    \toprule
    Method & Other HP settings \\
    \midrule
    \vae & train iters: [250,500] \\
    \rda & $\lambda$: [1e-1,1e-3,1e-5], iters: [20,30], inner iters: [20,30]  \\
    \svdd & LeakyRelu Slope: [1e-1, 1e-3], pretrain iters: [100,350],\\
    & pretrain lr: [1e-4], train iters: [250,500]]\\
    \rand  &  pretrain iters: [100], pretrain lr:[1e-4] adaptive sampling rate: [1.0]\\
    & train iters: [250,500], ens\_size = [5]\\
    \bottomrule
    \end{tabular}
    \label{tab:HP overview}
    \vspace{-0.2in}
\end{table}

\begin{table*}[ht]
\vspace{-0.15in}
    \small
    \centering
    \caption{\method architecture overview}
    \vspace{0.05in}
    \begin{tabular}{lr}
    \toprule
    List of HPs  & Settings \\
    \midrule
    BatchEnsemble num\_models & 8 (implicit ensemble over decay rate: [1.5,1.75,2,2.25,2.5,2.75,3,3.25]) \\
    num\_layers & 6 (implicit ensemble over AE-2, AE-4, AE-6,AE-8,AE-10,AE-12)  \\
    Train iterations & [250,500]\\
    Train Learning Rate & [1e-3,1e-4]\\
    Dropout rate & [0.0, 0.2] \\
    Weight decay & [0, 1e-5]\\
    \bottomrule
    \end{tabular}
    \label{tab:HPROBODoverview}
    \vspace{-0.2in}
\end{table*}

\subsubsection{Dataset Description}
\label{sssec:datadetails}
Similar to the experiment settings in our sensitivity analysis, we evaluate the baseline methods and \method on image data (MNIST, CIFAR10) as well as tabular data (Thyroid, Cardio and Lympho). For MNIST, we conduct three sets of experiments; each chooses digit `4',`5', or `8'
as the inlier class, respectively. For CIFAR10, we conduct two sets of experiments, with `airplane' (CIFAR10-0) and `automobile' (CIFAR10-1) as the inlier classes. For image data, we employ global contrast normalization to individual images. The inliers are assigned the label 0 and all classes other than the inlier class will be marked with 1, indicating the outlier class. We conduct all experiments under the Polluted setting, where we use all the inlier class points from  Pytorch's train data-split as label 0, and combine them with $10\%$ of points from the outlier classes within the train data-split. The tabular data, Thyroid, Cardio and Lympho, are downloaded from the ODDS repository (available at \url{http://odds.cs.stonybrook.edu/}, which contain $2.5\%$, $9.6\%$ and $4.1 \%$ outliers, respectively. Similar to image data, the inliers have label 0 and outliers have label 1. Prior to model training, we transform and scale each feature between zero and one using the MinMaxScaler.

\subsection{Additional Experiment Results}
\label{ssec:extra}

Fig. \ref{fig:mnist5examples} is shown to illustrate the regularization effect of parameter sharing in  \aes versus a vanilla AE. (See \ref{ssec:reg} for discussion.) 

\begin{figure}[!ht]
 \vspace{-0.15in}
	\centering
	\begin{tabular}{cc}
		\includegraphics[width=0.38\textwidth]{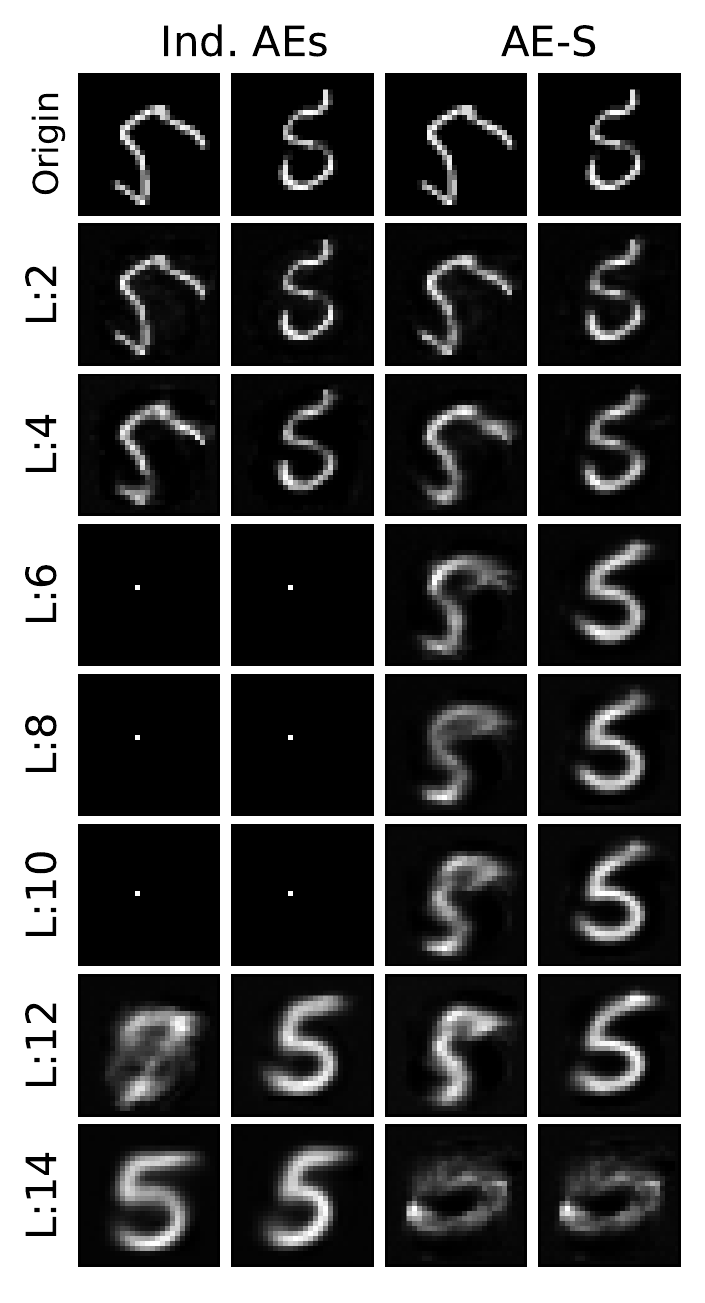} & \includegraphics[width=0.38\textwidth]{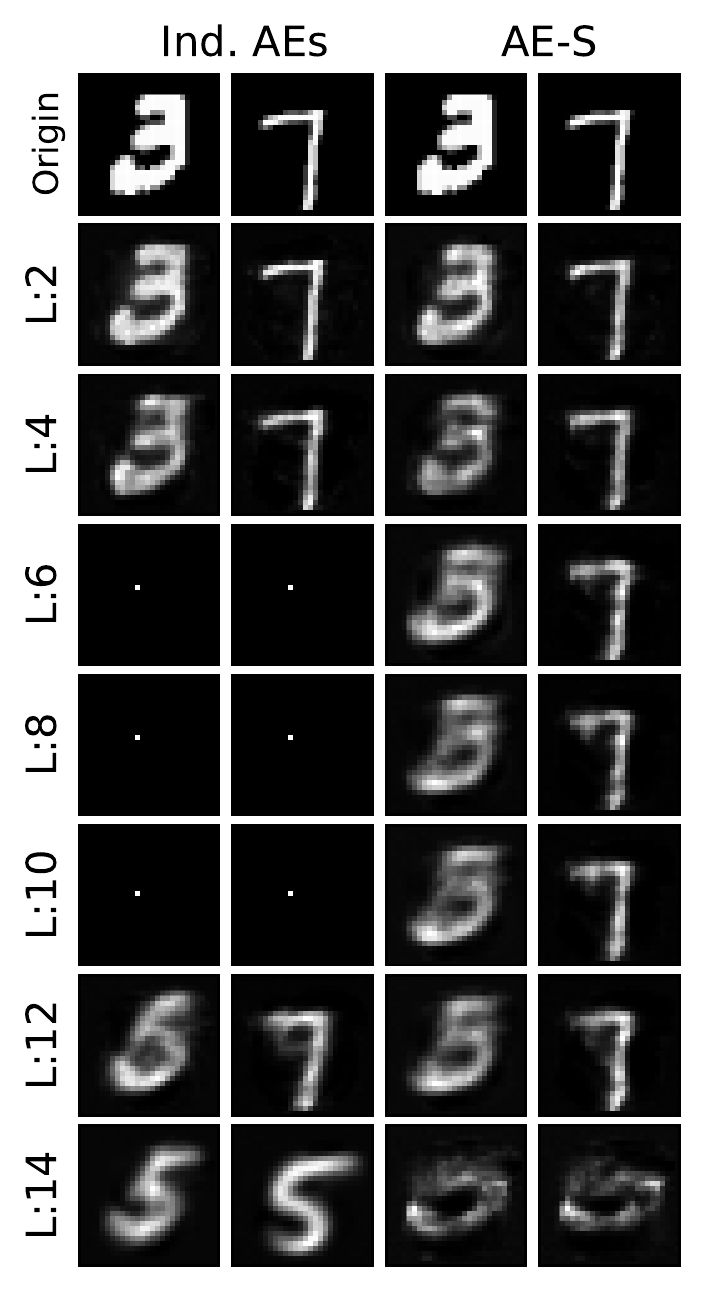} \\ 
\end{tabular}
 \vspace{-0.15in}
	\caption{\label{fig:mnist5examples} Left: Reconstructed \textit{inlier} class instances (MNIST digit `5'), generated by individual AEs vs AE-S structure in \method. Right: Reconstructed \textit{outlier} class instances (MNIST digit `3' and `7') by individual AEs versus AE-S structure. L denotes the number of layers.}
\end{figure}

Fig. \ref{fig:runtimevsauc} shows the running time (in log scale) vs. AUROC performance of OD methods (symbols) on datasets MNIST-4, MNIST-5, CIFAR-airplane, CIFAR-automobile, Thyroid and Lympho.

\begin{wrapfigure}{l}{\textwidth}
 \vspace{-0.1in}
	\centering
	\begin{tabular}{ccc}
		\includegraphics[width=0.3\textwidth]{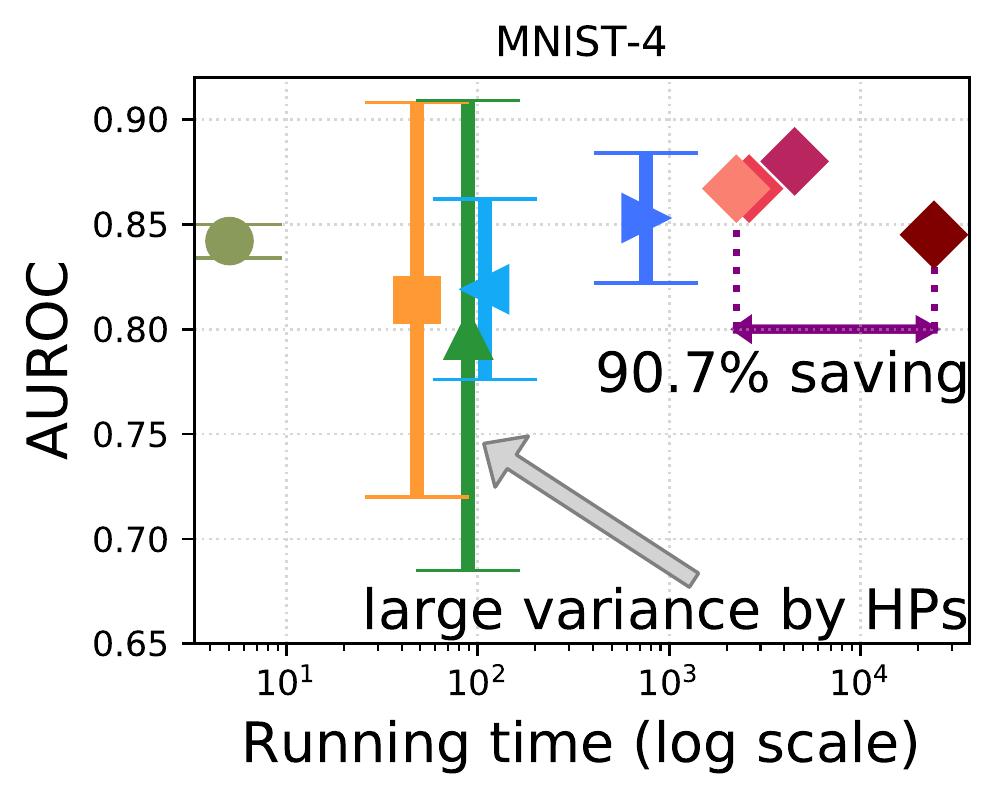} & 	\hspace{-0.08in}
		\includegraphics[width=0.3\textwidth]{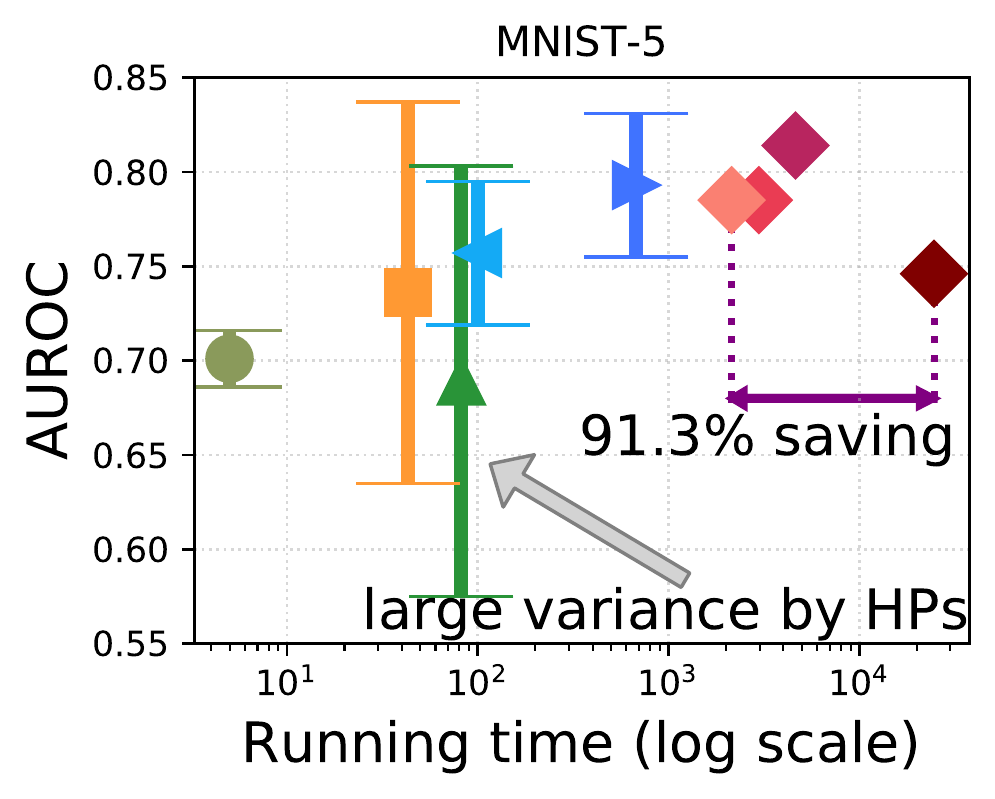} & 		\hspace{-0.08in}
		\includegraphics[width=0.3\textwidth]{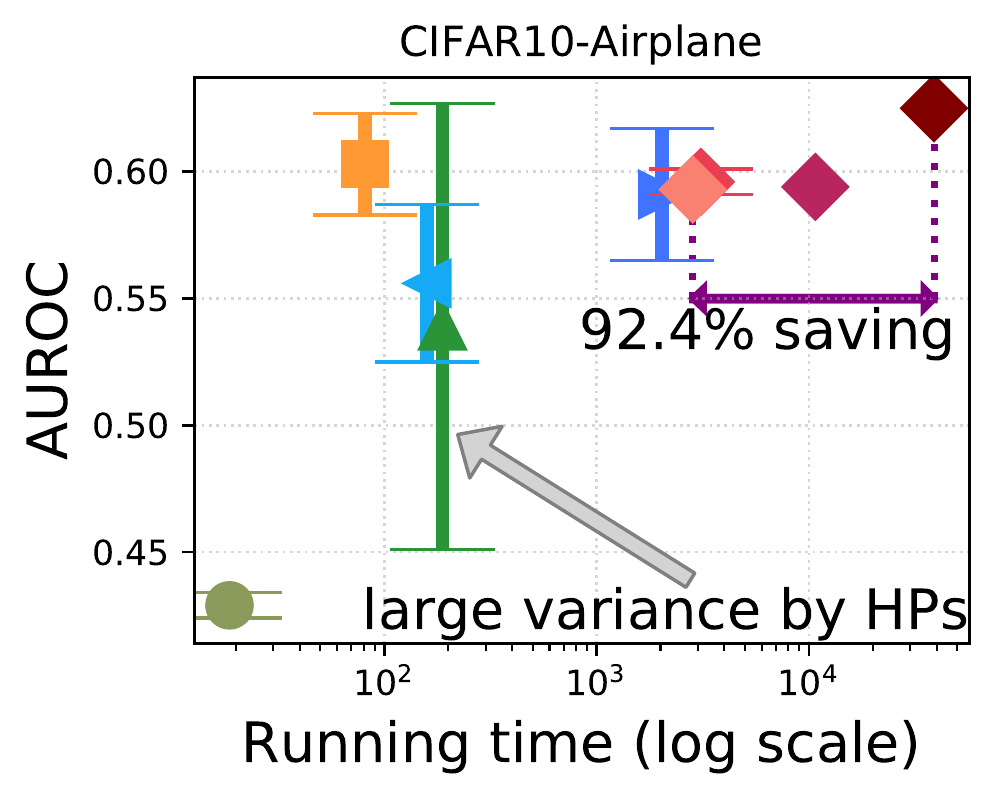} 	\\
		\includegraphics[width=0.3\textwidth]{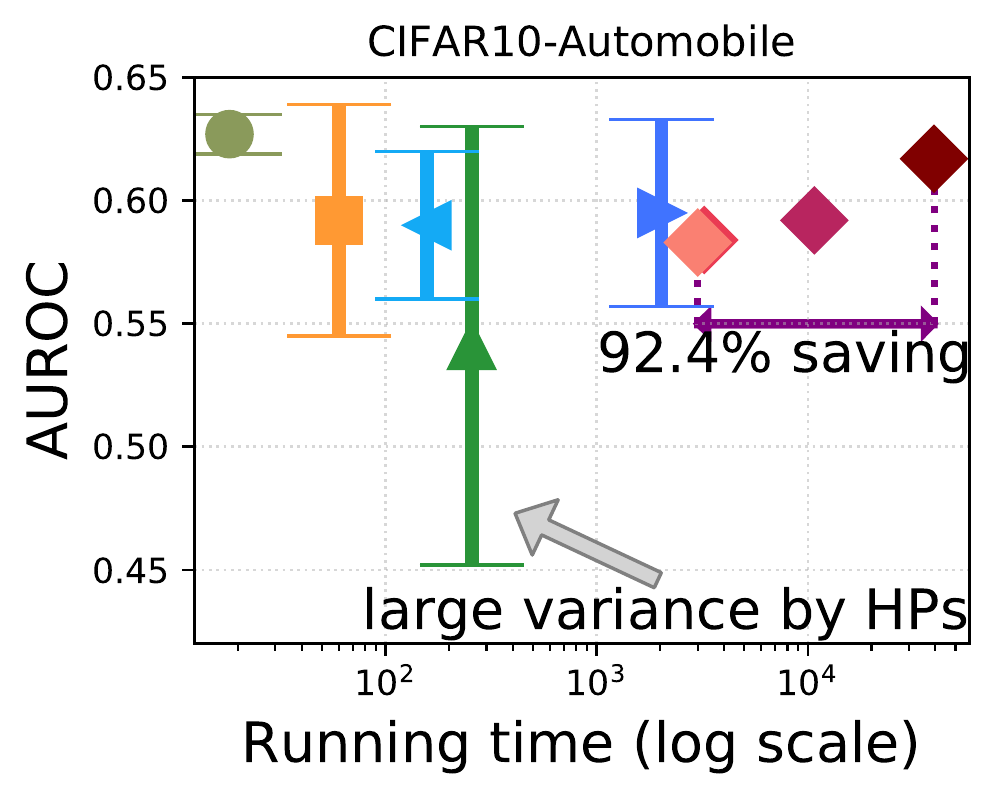} & 		\hspace{-0.08in}
		\includegraphics[width=0.3\textwidth]{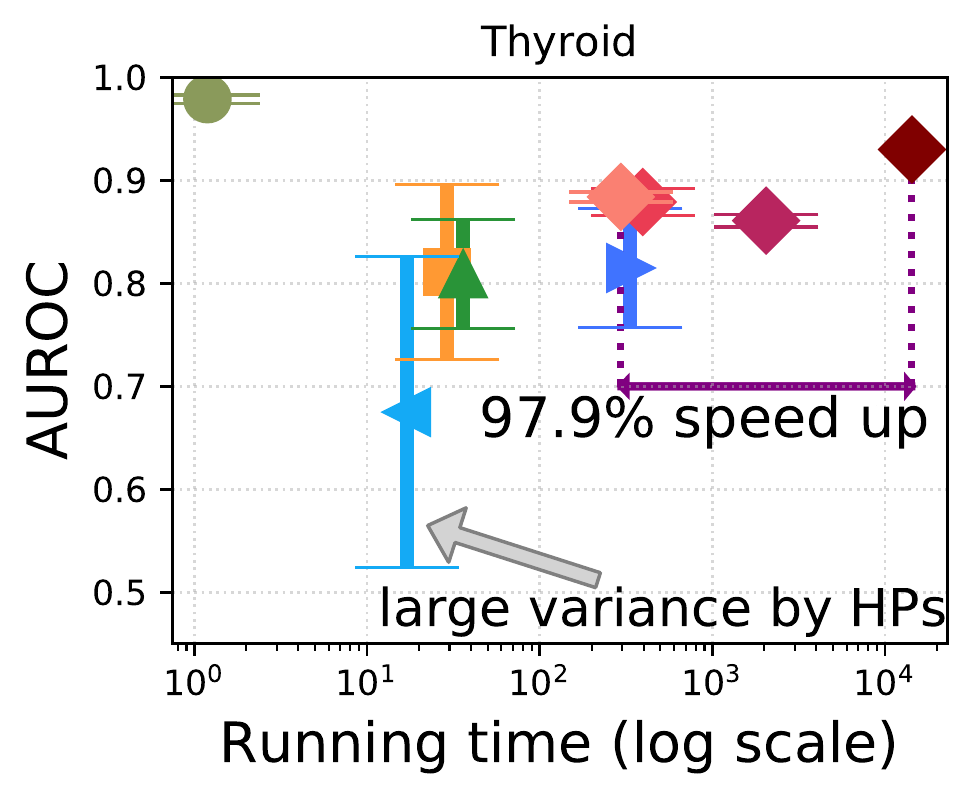} & 		\hspace{-0.08in}
		\includegraphics[width=0.3\textwidth]{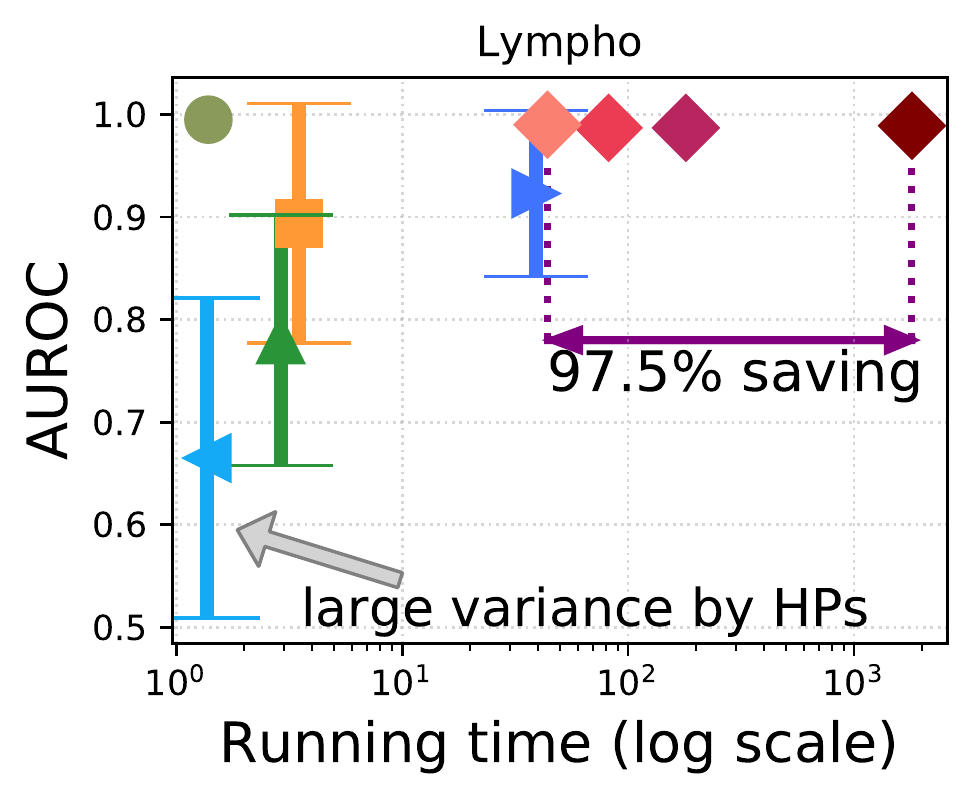} 	\\
		\multicolumn{3}{c}{	\hspace{-0.15in}
		\includegraphics[width=0.9\textwidth]{plots/legend} }
	\end{tabular}
	 \vspace{-0.1in}
	\caption{\label{fig:runtimevsauc}  Running time (in log scale) vs. AUROC of OD methods (symbols) on other datasets. Vertical bars depict one (1) stdev across HP config.s. \method often improves detection performance and importantly, provides robust low-variance performance. Sampling based \method reduces running time considerably with small difference in relative performance. 
	}
	 \vspace{-0.15in}
\end{wrapfigure}

\clearpage

\subsection{HP Sensitivity Analysis for Ensemble Models}
In this section, we want to show whether or not hyper-ensembles are robust to their own HPs: (1) the HP value ranges, and (2) the number of sub-models. We remark that the number of sub-models and HP ranges are directly related for a \textit{hyper}-ensemble, since expanding (or shrinking) the HP ranges result in more (or fewer) sub-models constituting the ensemble. Finally, we show how the proposed \method is robust to both the number of sub-models and varying HP value ranges.

\subsubsection{How do HP value ranges affect the ensemble?}
\label{sssec:HP-ranges}
Starting with the same HP settings as in Table \ref{tab:HP overview} for \vae, we expand, shrink or shift the HP ranges for the \vae models, and then measure the resulting accuracy and variance of \imethod, which assembles multiple {\vae}s. Table \ref{tab:HPrange-i} summarizes the various actions taken as compared to the original settings. For example, in Row 1, the range for the number of auto-encoders  is expanded to also include $[7,8,9]$ layers, or shifted to exhibit auto-encoders with $[6,7,8,9]$ layers, shifting from original $[2,3,4,5,6]$ layers. Other HPs are also altered in a similar fashion.

\begin{table*}[ht]
	\vspace{-0.05in}
	\small
	\centering
	\caption{Overview of \imethod with several HP ranges}
	\vspace{-0.02in}
	\begin{tabular}{lrr}
		\toprule
		List of HPs  & Original Setting & Actions \\
		\midrule
		Number of encoder layers & [2,3,4,5,6] & Expand:[2,3,4,5,6,7,8,9], Shift:[6,7,8,9], Shrink:[2,3,4] \\
		Train iterations & [250,500] & Shift\&Shrink:[1000], Expand:[250,500,1000]\\
		Decay rate & [1.5-3.25]  & Shrink:[1.5-2.0]  \\
		Train Learning Rate & [1e-3,1e-4] & Shift:[1e-2,1e-3], Expand:[1e-2,1e-3,1e-4]\\
		Dropout rate & [0.0, 0.2] & Shift:[0.2,0.5], Expand:[0.0,0.2,0.5]\\
		Weight decay & [0.0, 1e-5] & Shift: [1e-3,1e-4], Expand: [1e-3,1e-4,1e-5,0.0] \\
		\bottomrule
	\end{tabular}
	\label{tab:HPrange-i}
\end{table*}

We conduct our experiments on the same MNIST dataset with `4' being the inlier digit. For each  experiment, we alter a subset of the HPs from Table \ref{tab:HPrange-i} and record the AUROC across $3$ runs. We measure the mean and standard deviation of \imethod AUROC in comparison with \vae's.   Table \ref{tab:irobod-overall-result} provides the mean and variance of \imethod, evaluated on the performances over all these experimental runs, in comparison to training individual \vae. (Detailed results are in Table \ref{tab:HPrange-result-i}.)

Our results shows that \imethod produces more stable results in all settings than individually trained models. Moreover, it produces lower variance with respect to its own HPs ($2.8$) than that of the individual model results ($9.8$). Because of the lack of any prior knowledge, for many new models and architectures, finding a ``good'' HP range (a set of HPs that potentially contain the best HP for the task) can be hard. In that case, both the hyper-ensemble and an individual model may achieve less than satisfactory performance. However, hyper-ensemble is likely to achieve better stability to the range of HPs, while individual models are more sensitive when finding the optimal range is unreachable due to lack of validation.

\begin{table}[!h]
	\vspace{-0.0in}
	\setlength{\tabcolsep}{5pt}
	\caption{The altered HP ranges and AUROC results for \imethod and \vae. For example, ``Number of encoder layers:[6,7,8,9]'' means the values are shifted from [2,3,4,5,6] in the original experiment settings to [6,7,8,9], while the other HPs are the same as before as shown in Table \ref{tab:HPrange-i}.
	}
	\vspace{-0.1in}
	\hspace{-0.25in}
	\centering
	{{
			\begin{tabular}{l|r@{$\pm$}l | r@{$\pm$}l}
				\toprule
				Altered HP Ranges  & \multicolumn{2}{c}{\textbf{Mean\&Std.} (\imethod) }  & \multicolumn{2}{c}{ \textbf{Mean\&Std.} (\vae)}\\
				\midrule
				No Changes & 84.5&\textbf{0.0} & 81.4&9.4 \\
				Number of encoder layer: [2,3,4] & 86.1&\textbf{0.0} & 85.1&6.0 \\
				Number of encoder layer:[6,7,8,9] & 77.4&\textbf{0.1} & 75.1&7.1 \\
				Number of encoder layer: [2,3,4,5,6,7,8,9] &  82.1&\textbf{0.0} & 79.6&8.9 \\
				Train learning rate: [1e-2,1e-3] & 83.3&\textbf{0.2}      &  79.7&12.5 \\
				Train learning rate: [1e-2,1e-3,1e-4] & 83.2&\textbf{0.1}        & 79.4&12.3 \\
				Dropout rate: [0.2,0.5] & 83.6&\textbf{0.2}  & 80.7&9.5 \\
				Dropout rate: [0.0,0.2,0.5] & 84.1&\textbf{0.1}  & 81.1&9.6 \\
				Train iterations:[1000] & 84.5&\textbf{0.2} & 80.1&9.6 \\
				Train iterations:[250,500,1000] & 84.5&\textbf{0.1} & 81.2&9.0  \\
				Weight decay: [1e-3,1e-4] & 79.4&\textbf{0.1} & 77.9&3.8 \\
				Weight decay: [1e-3,1e-4,1e-5,0.0]   & 82.6&\textbf{0.1} & 80.2&8.2  \\
				Decay rate: [1.5,1.75,2.0] &  81.4&\textbf{0.1} & 78.6&13.4  \\
				\bottomrule                                                
			\end{tabular}
	}}
	\label{tab:HPrange-result-i}
\end{table}

\subsubsection{How does the number of sub-models affect the ensemble?}
\label{sssec:HP-numbers}
Next we investigate how the number of sub-models change the performance of various deep ensemble models. Here, we choose \svdd and \vae as the baseline ensembles, for which results under 764 and 6,912 individual models are reported in Table \ref{tab:HP-submodels}, conducted on MNIST-4 dataset. We subsample the sub-models among all the $764$ and $6,912$ models, with sizes equal to $[1,5,10,15,20,30,50,100,200,500]$. We conduct each subsampling $100$ times independently, and since we have $3$ experimental runs reported in the previous sections of the paper, we provide the results regarding detection accuracy and variances among $3\times 100$ experimental runs.

\begin{table}[!t]
	\caption{
		We define a grid of  values for each HP of our studied \svdd and \vae.
		With 4-to-8 different HPs each, the total number of configurations quickly grows to several hundreds.
	}
	\vspace{-0.1in}
	\hspace{-0.1in}
	\centering
	{\scalebox{0.7}{
			\begin{tabular}{l|lllllll}
				\hline
				\multicolumn{1}{|l|}{\textbf{Method}}             & \multicolumn{1}{l|}{\textbf{Hyperparameter}} & \multicolumn{1}{l|}{\textbf{Grid}}    & \multicolumn{1}{l|}{\textbf{\#values}} & \multicolumn{1}{l|}{\textbf{Method}}           & \multicolumn{1}{l|}{\textbf{Hyperparameter}} & \multicolumn{1}{l|}{\textbf{Grid}}    & \multicolumn{1}{l|}{\textbf{\#values}} \\ \hline
				\multicolumn{1}{|l|}{\multirow{4}{*}{\vae}} & n\_layers                                    & {[}2,3,4,5, 6,7,8,9{]}                           & \multicolumn{1}{l|}{8}                 & \multicolumn{1}{l|}{\multirow{8}{*}{\svdd}} & conv\_dim                                    & {[}8, 16, 32{]}                         & 3                                      \\
				\multicolumn{1}{|l|}{}                            & layer\_decay                                 & {[1.5,1.75,2,2.25,2.5,2.75,3,3.25]}                           & \multicolumn{1}{l|}{8}                 & \multicolumn{1}{l|}{}                          & fc\_dim                                      & {[}16, 32{]}                           & 2                                      \\
				\multicolumn{1}{|l|}{}                            & LR                                           & {[}1e-2, 1e-3, 1e-4{]}                 & \multicolumn{1}{l|}{3}                 & \multicolumn{1}{l|}{}                          & Relu\_slope                                  & {[}1e-1, 1e-3{]}                       & 2                                      \\
				\multicolumn{1}{|l|}{}                            & iter                                         & {[}200, 500, 1000{]}                    & \multicolumn{1}{l|}{3}                 & \multicolumn{1}{l|}{}                          & pretr\_iter                                  & {[}200, 350, 400{]}                     & 3                                      \\
				\multicolumn{1}{|l|}{}
				& \multicolumn{1}{l}{wght\_dc }                        &   [1e-3,1e-4,1e-5,0]       &   \multicolumn{1}{l|}{4}             & \multicolumn{1}{l|}{}                          & pretr\_LR                                    & {[}1e-4, 1e-5{]}                      & 2                                      \\
				\multicolumn{1}{|l|}{}        & Dropout                                   & {[}0.0, 0.2, 0.5{]}                  & \multicolumn{1}{l|}{3}                 & \multicolumn{1}{l|}{}                          & iter                                         & {[}100, 200, 250{]}                     & 3                                      \\ \cline{1-4}
			 	\multicolumn{1}{|l|}{}                            &                                     &  \multicolumn{1}{|l|}{Total \#models =}                       &  \multicolumn{1}{l|}{6,912}                 & \multicolumn{1}{l|}{}                          & LR                                           & {[}1e-4, 1e-5{]}                      & 2                                      \\ \cline{1-4}
				\multicolumn{1}{l}{}                            &                                &                          & \multicolumn{1}{l|}{}                 & \multicolumn{1}{l|}{}                          & wght\_dc                                     & {[}1e-5, 1e-6{]}                      & 2                                      \\ \cline{5-8} 
				\multicolumn{1}{l}{}                            &                                            &                       &                                      & \multicolumn{1}{|l}{}                          & \multicolumn{1}{l|}{}                        & \multicolumn{1}{|l|}{Total \#models =} & \multicolumn{1}{l|}{864}               \\ 
				\cline{5-8}
			\end{tabular}
	}
}
	\vspace{-0.2in}
	\label{tab:HP-submodels}
\end{table}

Fig. \ref{fig:submodelvsauc} shows the AUROC corresponding to different number of sub-models (for \svdd we have both Polluted (left) and Clean (middle) settings, and for \vae we have Polluted (right) setting only). Notice that when the number of sub-models is less than 20, the overall performance variance is relatively larger, where the ensemble performance is similar to an individual model prediction. AUROC quickly stabilizes and the variance shrinks as the number of sub-models becomes larger than 20 –especially for the \svdd's Clean setting and \vae's Polluted setting– with little difference beyond. These results suggest that ensembles are not sensitive to the number of sub-models beyond a certain size, where the larger the number, the more stable is the performance.  



\subsubsection{HP Sensitivity Analysis: \method}
\label{sssec:HP-robod}
In this section, we perform the sensitivity analysis to both HP value ranges as well as the number of sub-models (since they are correlated) for our proposed \method. We extend, shrink or shift the number of layers and number of BatchEnsemble models within the \aes structure, which corresponds to changing the HP-ranges of NN widths and depths. We also recognize the additional training HPs such as train iterations, learning rate, dropout rate, weight decay, etc. Table \ref{tab:ROBODrange} summarizes the HP-ranges we used to conduct our HP-sensitivity analysis based on Table \ref{tab:HPROBODoverview}. These additional experiments are on the Cardio dataset. For each set of HP configurations, we repeat $3$  experimental runs and summarizes mean and variance of \method's AUROC. The results among different hyper-ensembles (for both HP ranges and number of sub-models) are given in the following two tables, Table \ref{tab:HPrange-result} (which alters the \aes structure HPs) and Table \ref{tab:HPrange2-result} (which alters the other training HPs).

Our results on \method match with the two observations in the previous subsections, in that \method, the sped-up version of the \imethod hyper-ensemble, provides stable results with the varying (1)  number of sub-models, and (2) HP value ranges. Moreover, the mean performance and standard deviation of all \method experiments listed in Tables \ref{tab:HPrange-result} and  \ref{tab:HPrange2-result} are considerably more competitive than all the other benchmarked models we studied, while \method yields smaller variance to its own HP configurations than many banchmarked models (Table \ref{tab:irobod-comparison}).

Our analyses provide practical insights on how to reduce an ensemble model's sensitivity to their HP settings. As long as time and resources permit, one should employ as many number of sub-models as possible, and expand the HP value ranges under a fine grid. In contrast, finding a single set of optimal/good HPs for an individual OD model is almost infeasible for the unsupervised setting. 

\begin{table*}[!t]
	\small
	\centering
	\caption{\method HP-ranges overview. Note that num\_models corresponds to implicit ensemble over decay rate [1.5,1.75,2,2.25,2.5,2.75,3,3.25] (see Table \ref{tab:HPROBODoverview}), and num\_layers corresponds to implicit ensemble over \aei; e.g.  num\_layers=6 assembles over AE-2, AE-4, AE-6, AE-8, AE-10, and AE-12.}
	\vspace{0.05in}
	\begin{tabular}{lrr}
		\toprule
		List of HPs  &  Original Settings & Actions \\
		\midrule
		BatchEnsemble num\_models & 8  & Shrink:[4] Shrink:[5] Shrink:[6] Shrink:[7]\\
		num\_layers & 6  & Shrink:[4] Expand:[8] \\
		Train iterations & [250,500] & Shift:[500,1000] Expand:[250,500,1000] Shrink: [250]\\
		Train Learning Rate & [1e-3,1e-4] & Shift:[1e-1,1e-2] Expand: [1e-1,1e-2,1e-3,1e-4]\\
		Dropout rate & [0.0, 0.2]  & Shrink\&Shift:[0.5] Expand: [0.0,0.2,0.5] \\
		Weight decay & [0, 1e-5] & Expand: [1e-4,1e-5,0] Shrink: [0]\\
		\bottomrule
	\end{tabular}
	\label{tab:ROBODrange}
	\vspace{0.1in}
\end{table*}

\begin{table}[!h]
	\vspace{0.1in}
	\setlength{\tabcolsep}{5pt}
	\caption{Results of \method over different \aes structures. With different num\_models and num\_layers, various number of AE models are implicitly assembled. We report the mean AUROC and standard deviation across $3$ runs, along with the number of (implicit) sub-models in parenthesis.
	}
	\vspace{-0.06in}
	\centering
	{{
			\begin{tabular}{lrrrr}
				\toprule
				\multicolumn{4}{c}{\textbf{Mean\&Std.} for different \aes structures and the number of sub-models} \\
				\midrule
				& num\_models:4 & num\_models:5 & num\_models:6 & num\_models:7 \\ 
				num\_layers:4 & 92.4$\pm$0.6 (128)   &  93.3$\pm$0.3 (160)  & 93.1$\pm$0.4 (192)  &  93.3$\pm$0.2 (224)  \\ 
				num\_layers:6 & 93.6$\pm$0.4 (192) & 93.6$\pm$0.1 (240)  &  93.5$\pm$0.1 (288) & 93.8$\pm$0.3 (336)   \\ 
				num\_layers:8 & 93.8$\pm$0.1 (256)  & 93.7$\pm$0.1 (320)  &  93.5$\pm$0.1 (384)  &  93.8$\pm$0.1 (448) \\ \hline
			\end{tabular}
	}}
	\label{tab:HPrange-result}
\end{table}

\begin{table}[!h]
	\vspace{0.1in}
	\setlength{\tabcolsep}{5pt}
	\caption{The altered HPs and results for \method. For example, ``Train iterations: [250,500,1000]'' means that the training iterations are expanded from the original experiment settings (see Table \ref{tab:ROBODrange}) to include the additional value [1000], while the \aes HPs are the same as before (BatchEnsemble num\_models equals to 8, num\_layers equals to 6). Mean and Std. of AUROC over 3 runs are reported.
	}
	\vspace{-0.08in}
	\hspace{-0.25in}
	\centering
	{{
			\begin{tabular}{l|r@{$\pm$}l}
				\toprule
				Altered HP Ranges  & \multicolumn{2}{c}{\textbf{Mean\&Std.} (\method) }  \\
				\midrule
				No Changes (384 submodels) & 93.5&0.1  \\
				Train iterations: [250,500,1000] (576 submodels) & 93.7&0.1 \\
				Train iterations: [250] (191 submodels) & 94.0&0.2\\
				Train Learning Rate: [1e-1,1e-2] (384 submodels) &93.7&0.2 \\
				Train Learning Rate: [1e-1,1e-2,1e-3,1e-4] (576 submodels) &93.8&0.0\\
				Dropout Rate: [0.5] (191 submodels)& 93.7&0.1 \\
				Dropout Rate: [0.0,0.2,0.5] (576 submodels) & 93.6&0.2\\
				Weight Decay: [1e-4,1e-5,0.0] (576 submodels)& 94.0&0.0\\
				Weight Decay: [0.0] (191 submodels) & 93.7&0.3\\
				\bottomrule                                                
			\end{tabular}
	}}
	\label{tab:HPrange2-result}
\end{table}




\end{document}